\documentclass[accepted]{article}

\usepackage{microtype}
\usepackage{graphicx}
\usepackage{booktabs} 

\usepackage{hyperref}

\usepackage{icml2023}

\usepackage{amsmath}
\usepackage{amssymb}
\usepackage{mathtools}
\usepackage{amsthm}

\usepackage[capitalize,noabbrev]{cleveref}

\renewcommand{\epsilon}{\varepsilon}
\usepackage{dsfont}
\usepackage{graphicx}
\usepackage{subcaption}
\usepackage{sidecap}
\usepackage{paralist}

\definecolor{myblue}{HTML}{1f77b4}
\definecolor{myorange}{HTML}{ff7f0e}
\definecolor{mygreen}{HTML}{2ca02c}
\definecolor{myred}{HTML}{d62728}

\DeclareMathOperator*{\argmax}{argmax}

\newcommand\colorrule[3][black]{\textcolor{#1}{\rule{#2}{#3}}}

\usepackage{commenting}
\usepackage{fancyvrb}
\VerbatimFootnotes

\usepackage{soul}

\newcommand\plotscale{0.8}

\theoremstyle{plain}

\theoremstyle{definition}

\theoremstyle{remark}

\usepackage[textsize=tiny]{todonotes}

\icmltitlerunning{Internally Rewarded Reinforcement Learning}

\begin{document}

\twocolumn[
\icmltitle{Internally Rewarded Reinforcement Learning}



\icmlsetsymbol{equal}{*}

\begin{icmlauthorlist}
\icmlauthor{Mengdi Li}{equal,wtm}
\icmlauthor{Xufeng Zhao}{equal,wtm}
\icmlauthor{Jae Hee Lee}{wtm}
\icmlauthor{Cornelius Weber}{wtm}
\icmlauthor{Stefan Wermter}{wtm}
\end{icmlauthorlist}

\icmlaffiliation{wtm}{Knowledge Technology Group, Department of Informatics, University of Hamburg, Hamburg, Germany}
\icmlcorrespondingauthor{Mengdi Li}{mli@informatik.uni-hamburg.de}

\icmlkeywords{Reinforcement Learning}

\vskip 0.3in
]



\printAffiliationsAndNotice{\icmlEqualContribution} 

\setcounter{footnote}{1}

\begin{abstract}
We study a class of reinforcement learning problems where the reward signals for policy learning are generated by an internal reward model that is dependent on and jointly optimized with the policy. This interdependence between the policy and the reward model leads to an unstable learning process because reward signals from an immature reward model are noisy and impede policy learning, and conversely, an under-optimized policy impedes reward estimation learning. We call this learning setting \emph{Internally Rewarded Reinforcement Learning} (IRRL) as the reward is not provided directly by the environment but \emph{internally} by a reward model. In this paper, we formally formulate IRRL and present a class of problems that belong to IRRL. We theoretically derive and empirically analyze the effect of the reward function in IRRL and based on these analyses propose the clipped linear reward function. Experimental results show that the proposed reward function can consistently stabilize the training process by reducing the impact of reward noise, which leads to faster convergence and higher performance compared with baselines in diverse tasks.\footnote{Project page: https://ir-rl.github.io/}
\end{abstract}

\section{Introduction}
\begin{figure}
    \centering%
    \includegraphics[width=\linewidth]{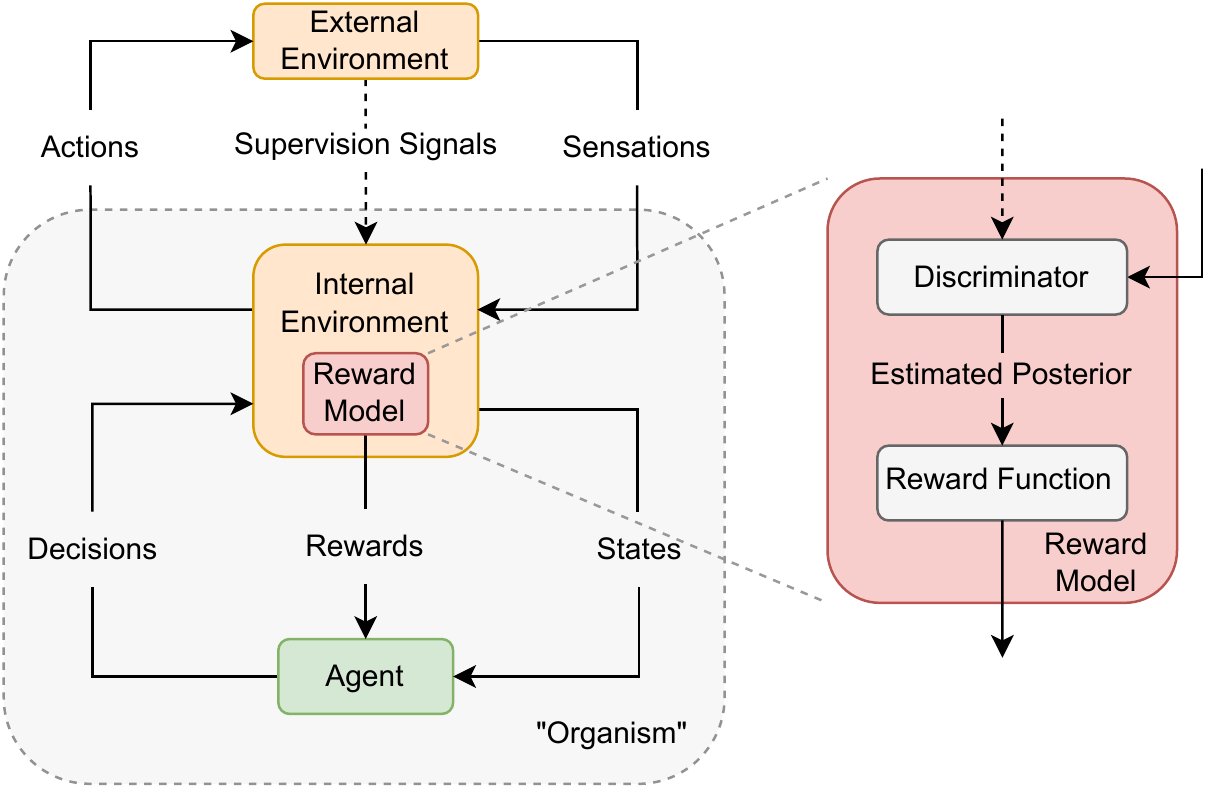}
    \caption{\textbf{Left:} The agent-environment interaction loop of IRRL. This diagram is based on the scheme of intrinsically motivated RL~\citep{Singh2004IntrinsicallyMotivatedReinforcement} with an optional path of supervision signals, which reflects an extrinsic reward. \textbf{Right:} The internal reward model consists of a \emph{discriminator}, which estimates a posterior probability of correct discrimination given sensations and supervision signals from the external environment, 
    and a \emph{reward function}, which produces rewards by processing the posterior probability. }
    \label{fig:elaborated-agent-environement-interaction-with-sl-and-noisy-rewards}
\end{figure}

Rewards are essential for animals and artificial agents to learn by exploration in an environment. In the brain, reward signals are emitted by specific neurons as a consequence of the processing of external stimuli~\cite{Olds1954PositiveReinforcementRroduced,Schultz2015NeuronalRewardDecision}. For instance, when a child receives words of praise from the parents as feedback for exhibiting appropriate behavior, the rewards obtained are contingent upon the child's individual understanding of the words. In some cases, the child may misunderstand the praise as criticism, thus wrongly obtaining a negative reward and impeding its behavior learning. An elaborated view of the standard agent-environment interaction formulation~\citep{suttonb98} of reinforcement learning (RL) demonstrates this mechanism~\citep{Singh2004IntrinsicallyMotivatedReinforcement}. This framework separates the environment into an \emph{external environment}, which provides external stimuli (e.g., a word of praise from the parents), and an \emph{internal environment}, which is in the same ``organism" with the agent and contains a reward model\footnote{We adopt the term ``reward model" instead of using ``critic", as used by \citeauthor{Singh2004IntrinsicallyMotivatedReinforcement} (\citeyear{Singh2004IntrinsicallyMotivatedReinforcement}), to prevent confusion with the term ``critic" in actor-critic algorithms \cite{KondaT99ActorCritic}. } that 
processes external stimuli and produces reward signals (cf.~Fig~\ref{fig:elaborated-agent-environement-interaction-with-sl-and-noisy-rewards} left panel).

In this work, we focus on situations where the reward is determined by both external stimuli and the state of a sophisticated and evolutionary internal environment that produces either task-relevant rewards 
\cite{mnihhgk14,bamk14,liwklz0w21,samrudhdhicj22}
or task-agnostic rewards 
\cite{Gregor17VariationalIntrinsic,Strouse2022LearningMoreSkills},
and we use the term \textit{Internally Rewarded Reinforcement Learning} (IRRL) to refer to the learning problem in these situations (cf. some IRRL examples in Fig.~\ref{fig:irrl-examples}). 

In IRRL, the policy of the agent is trained by RL, and the reward model of the internal environment is simultaneously trained either in a self-supervised learning (SSL) manner by directly using the sensations from the external environment~\citep{pathak2017curiosity,Gregor17VariationalIntrinsic,Eysenbach2019DiversityAllYou,Strouse2022LearningMoreSkills}, or in a supervised learning (SL) manner by using extra human-annotated task-relevant signals~\citep{mnihhgk14,yull17,tan0gz020,liwklz0w21, ChristianoLBMLA17deepRLfromHuman, Ouyang0JAWMZASR22}. 
The reward model provides reward signals for training a policy that, in return, controls the collection of the trajectories for the reward model. These scenarios have become prevalent with increased interest in integrating the capability of high-level prediction and low-level control of behaviors into a single model in the realms of attention mechanisms~\citep{mnihhgk14,bamk14,yull17,liylzwx17,samrudhdhicj22}, embodied agents~\citep{GordonKRRFF18,YangRXCCPB19}, robotics~\citep{lakomkinzwmw18,liwklz0w21}, unsupervised RL~\citep{Gregor17VariationalIntrinsic, Eysenbach2019DiversityAllYou, Strouse2022LearningMoreSkills}, and reinforcement learning from human feedback (RLHF)~\citep{ChristianoLBMLA17deepRLfromHuman, Ouyang0JAWMZASR22}. 

The role of the reward model depends on the target task. In the task of digit recognition with hard attention (see Fig.~\ref{fig:task-vis-digit-recognition}), for example, the reward model assesses the certainty of performing correct digit classification. In the unsupervised skill discovery task (see Fig.~\ref{fig:task-vis-skill-discovery}), however, the reward model works as an intrinsic motivation system to evaluate the novelty of generated skills. The reward model consists of a discriminator and a reward function, as shown in the right panel of Fig.~\ref{fig:elaborated-agent-environement-interaction-with-sl-and-noisy-rewards}. The discriminator estimates the posterior probability of the target label provided by supervision signals or sensations. 
Given the posterior, the reward function produces rewards for the behavior learning of the agent. 

\begin{figure}
    \centering%
    \includegraphics[width=0.5\linewidth]{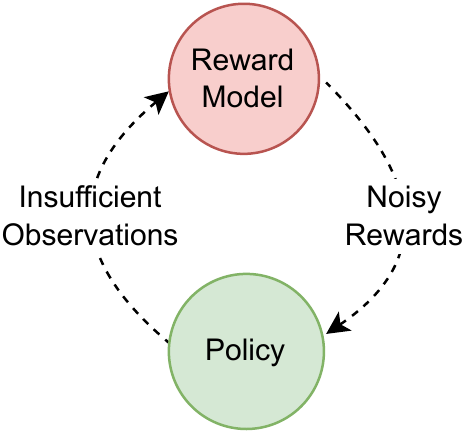}
    \caption{Simultaneous optimization between the policy of the agent and the reward model is challenging because an under-optimized reward model yields noisy rewards, and in turn, an immature policy yields insufficient observations, which leads to an unstable training loop. }
    \label{fig:unstable-training-loop}
\end{figure}

Simultaneous optimization between the policy and the discriminator in IRRL is however non-trivial because of the unstable training loop where neither of them can learn efficiently (see~Fig.~\ref{fig:unstable-training-loop}). 
In this work, we seek to solve this issue by reducing the impact of reward noise, which is challenging due to the unavailability of an oracle discriminator whose posterior probability can reflect the information sufficiency for discrimination.
We theoretically formulate IRRL to explicitly analyze the noisy reward issue
and characterize the distribution of the noise empirically by approximating 
the oracle discriminator with the discriminator of a converged model. Based on our formulation and empirical results, we demonstrate the effect of the reward function in reducing the bias of the estimated reward and the variance of the reward noise,
and propose a simple yet effective reward function that stabilizes the training process. 

We present extensive experimental results on IRRL tasks with task-relevant rewards (i.e., visual hard attention, and robotic active vision), or tasks with task-agnostic rewards (i.e., unsupervised skill discovery). The results suggest that our proposed reward function consistently improves the stability and the speed of training, and achieves better performance than the baselines on all the tasks. In particular, on the skill discovery task, our approach with the simple reward function achieves the same performance as the state-of-the-art sophisticated ensemble-based Bayesian method by~\citet{Strouse2022LearningMoreSkills} but without using ensembles. We further demonstrate that the superiority of the proposed reward function is due to its effectiveness in noise reduction, which is in line with our theoretical analysis.
The contributions of this paper are summarized as follows:
\begin{enumerate}
    \item We formulate a class of RL problems as IRRL, and formulate the inherent issues of \textit{noisy rewards} that leads to an unstable training loop in IRRL. 
    \item We empirically characterize the noise in the discriminator and derive the effect of the reward function in reducing the bias of the estimated reward and the variance of the reward noise stemming from an underdeveloped discriminator.
    \item We propose a simple yet effective reward function, the \emph{clipped linear reward function}, which consistently stabilizes the training process and achieves faster convergence speed and higher performance on diverse IRRL tasks.
\end{enumerate}

\section{Related Work}
The RL process is notoriously unstable. Previous work has studied various techniques to stabilize training, such as reducing the bias and variance of gradient estimation for policy gradient methods~\citep{Greensmith2004VarianceReductionTechniques, Schulman2015HighDimensionalContinuousControl}, and value estimation for value-based methods ~\citep{vanHasselt2016DeepReinforcementLearning}. As another factor impacting RL training, reward noise that stems from various sources, e.g., sensors on robots, and adversarial attacks, is attracting attention because of the growing interest in applying RL to more realistic and complicated tasks ~\citep{Huang2017AdversarialAttacksNeural,Everitt2017ReinforcementLearningCorrupted, Wang2020ReinforcementLearningPerturbeda}. In cases where the noise directly resides in the reward, both policy gradient and value-based RL methods suffer. \citet{Everitt2017ReinforcementLearningCorrupted} and \citet{Wang2020ReinforcementLearningPerturbeda} formulate RL with corrupted rewards and partially address the issue for cases with extra knowledge about the noise. Unlike the noise caused by reward corruption, the noise in IRRL comes from a discriminator and is subject to the learning process,
so their approaches are not directly applicable to our scenarios in terms of both formulation and experimental emulation.

The issues of unstable training in IRRL have been mentioned in the literature, but they have not been systematically studied. Some works~\citep{mnihhgk14,bamk14,liylzwx17} ignore the impact of the unstable training loop at the expense of the training speed and the performance of the final model. Other works resort to elaborated training strategies, e.g., staged training~\citep{GordonKRRFF18, YangRXCCPB19, LWBW22}, curriculum training~\citep{DasDGLPB18,liwklz0w21}, imitation learning~\citep{tan0gz020,samrudhdhicj22}, or task-specific reward shaping~\citep{DengGGZ0021}. However, extra efforts such as data collection or human ingenuity are needed in these methods.

\citet{Strouse2022LearningMoreSkills} study the pessimistic exploration problem in the context of unsupervised skill discovery (cf.~Fig.~\ref{fig:task-vis-skill-discovery}) where a skill discriminator is used to generate rewards. As the skill discriminator is subject to noise, this issue can be seen as a consequence of the unstable training loop under the framework of IRRL. Similar to our work, they also resort to modifying the reward function. They propose to train an ensemble of discriminators and reward the policy with their disagreement. Experimental results suggest that the proposed disagreement-based reward lets the agent learn more skills through optimistic exploration. However, this method introduces more model parameters and hyper-parameters than baseline methods that are not based on ensembles. 
In this paper, we consider the issue in a more general context including but not limited to unsupervised skill discovery, and manage the issue in a more simple and efficient way.

\section{Internally Rewarded RL}
\label{sec:related_work_internally_rewarded_rl}

\begin{figure*}[t]
    \subcaptionbox{%
      \textbf{Hard attention for digit recognition on the Cluttered MNIST dataset} \cite{mnihhgk14}. A small glimpse (the squares) controlled by an attention policy sequentially changes its location to collect information for recognizing the digit. During training, the reward model is expected to produce rewards that reflect the sufficiency of information collected by the attention policy, and in turn, the policy is expected to attend to informative regions, i.e., pixels of the digit, to collect information for the classifier to learn digit recognition. The starting and stopping glimpses are represented by yellow and red boxes respectively. The green line indicates the positions of intermediate glimpses.\label{fig:task-vis-digit-recognition}%
    }[.34\textwidth]{%
      \hfill\includegraphics[width=.14\textwidth]{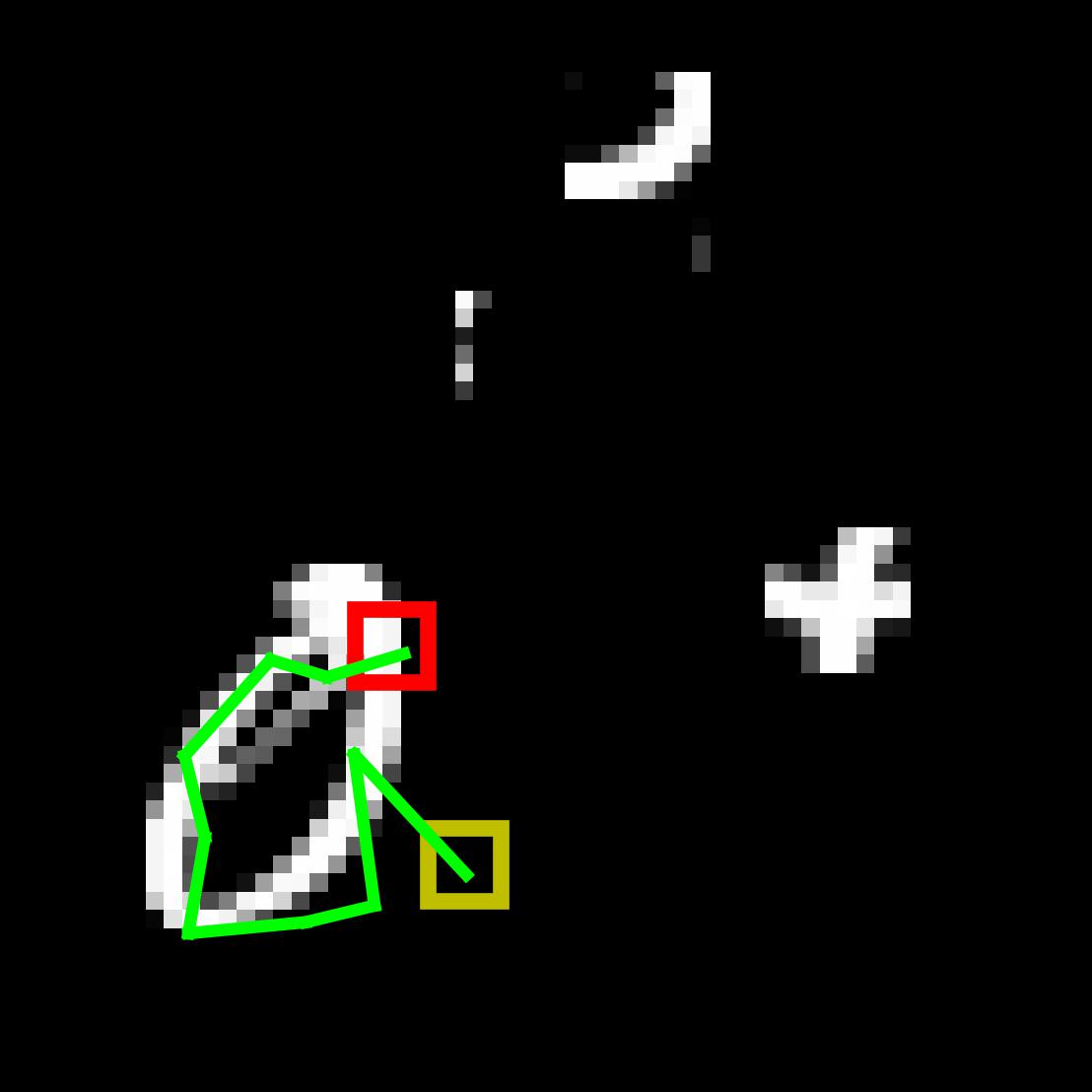}\hfill\includegraphics[width=.14\textwidth]{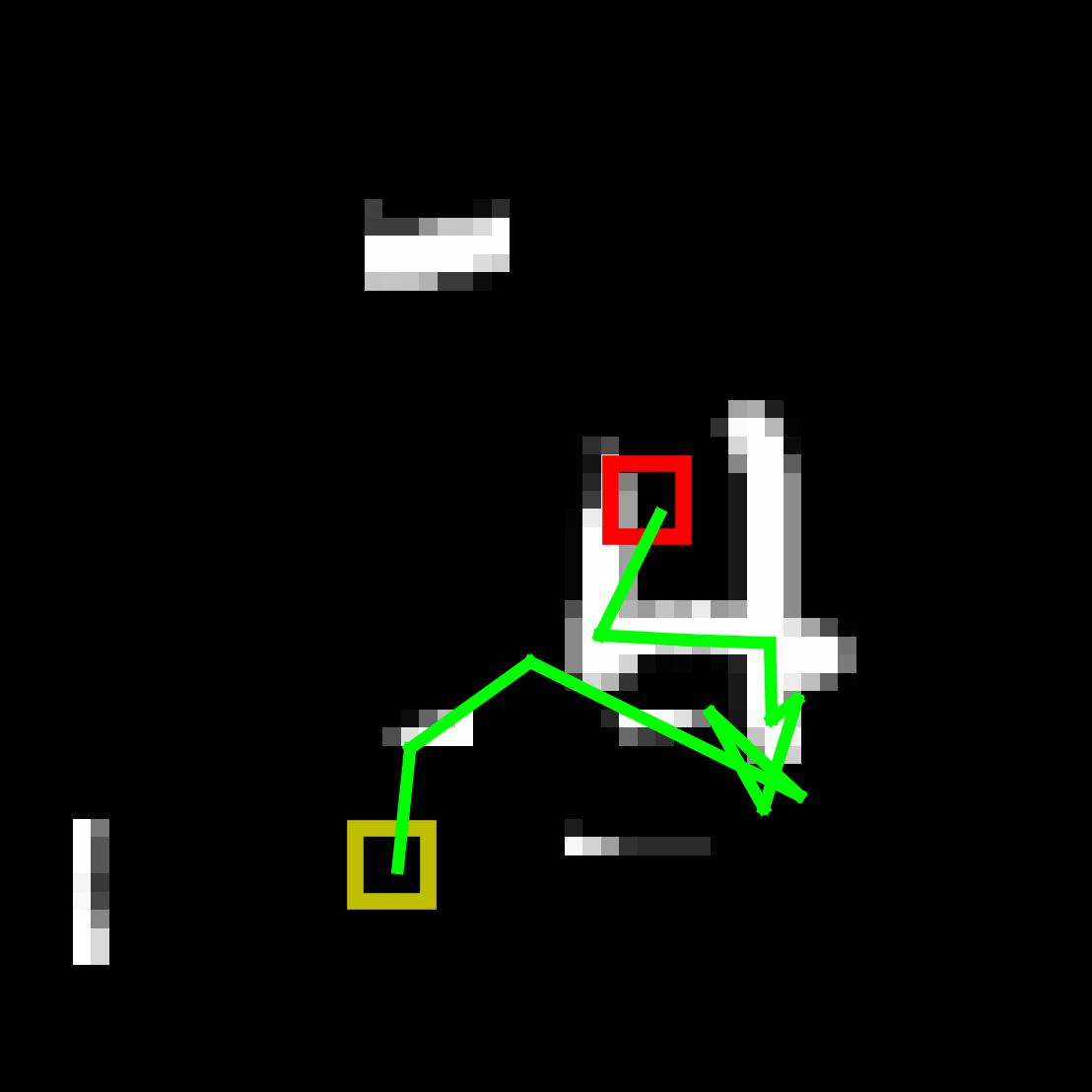}\hfill\phantom{}%
    }%
    \hfill%
    \subcaptionbox{\textbf{Unsupervised skill discovery in a four-room environment} \cite{Strouse2022LearningMoreSkills}. An agent spawned at the top-left corner is expected to learn a navigation policy that performs distinguishable skills without using any extrinsic rewards. In this task, a skill is represented by the final state of a trajectory. During training, the agent generates a trajectory conditioned on a randomly sampled skill label, and a discriminator estimates the posterior probability of the trajectory being the target skill, based on which the reward is produced. The policy and the discriminator are optimized simultaneously.\label{fig:task-vis-skill-discovery}}[0.3\textwidth]{\includegraphics[width=0.14\textwidth]{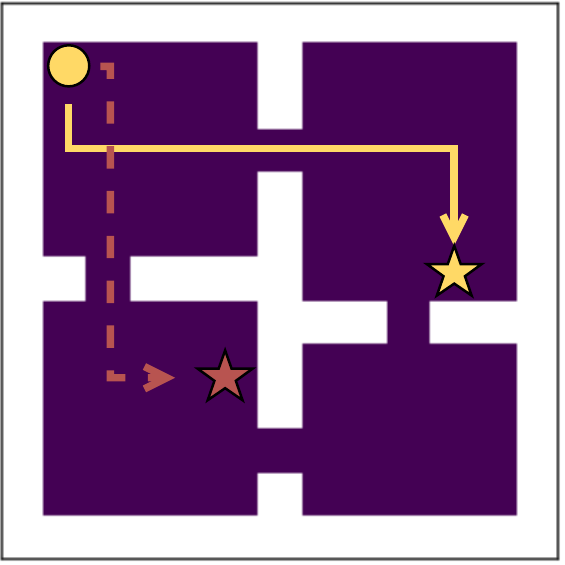}}%
    \hfill
    \subcaptionbox{%
      \textbf{Robotic object counting in occlusion scenarios}. A humanoid robot is trained to learn a locomotion policy to explore occluded space by rotating around the table and to terminate exploration to achieve efficient counting of specified objects, e.g., \textit{small\_blue\_cube}. The robot performs the task solely based on its egocentric RGB view. During training, the policy uses the reward that is produced by a reward model containing an object counter, which is simultaneously updated with the policy. Similar to the task of hard attention, the reward should be able to evaluate the information sufficiency of observations for correct object counting.\label{fig:task-vis-object-counting}%
    }[.32\textwidth]{%
      \hfill\includegraphics[width=.14\textwidth]{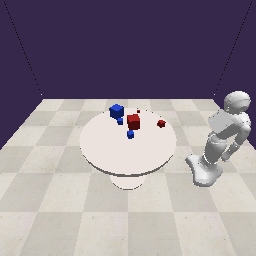}\hfill\includegraphics[width=.14\textwidth]{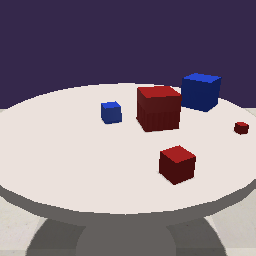}\hfill\phantom{}%
    }
    \caption{Example tasks of IRRL\label{fig:irrl-examples}}
\end{figure*}

We formulate the policy learning of IRRL as a Markov decision process $\mathcal{M} = \langle \mathcal{S}, \mathcal{A}, p_\text{E}, \rho, r, \gamma \rangle$, where, $\mathcal{S}$ is the state space, $\mathcal{A}$ the action space, $p_\text{E}: \mathcal{S} \times \mathcal{A} \times \mathcal{S} \rightarrow \mathbb{R}$ the state transition probability, $\rho: \mathcal{S} \rightarrow \mathbb{R}$ the distribution of the initial state, $r: \mathcal{S} \times \mathcal{A} \rightarrow \mathbb{R}$ the reward on each transition, and $\gamma \in (0,1)$ a discount factor.

Different from conventional RL settings, where reward $r$ depends exclusively on the \emph{external} environment, in IRRL reward $r$ is determined by a \emph{reward model}, which resides in the \emph{internal} environments and interprets the supervision signals from the external environment to generate internal rewards (cf.~Fig.~\ref{fig:elaborated-agent-environement-interaction-with-sl-and-noisy-rewards}). Here, we assume that the external environment, hence the observations an agent is making, is caused by a label $y$ sampled from a prior distribution $p(y)$. The reward model depends on a trainable \emph{discriminator} $q_\phi$ parameterized with $\phi$. Given a trajectory
\small
$\tau \in (\mathcal{S} \times \mathcal{A})^{n}$
\normalsize
($n \in \mathbb{N}$ is the trajectory length) sampled from a policy $\pi_\theta$ parameterized with $\theta$, the discriminator $q_\phi(y \mid \tau)$ computes the probability of label $y$ being the cause of trajectory $\tau$.\footnote{To simplify notations, we use lower-case letters (e.g., $y$) to both represent random variables and their realizations if the distinction is clear from the context. Similarly, we use $p(y)$ to both represent the distribution of $y$ and the probability of $y$ if the context is clear.}

Many existing works, which have been studied independently before, can be categorized as instances of IRRL.
In the subsequent discussion, we present three lines of existing works as concrete examples of IRRL:

\begin{asparaenum}
    \item \textbf{Hard attention}. Hard attention mechanism~\citep{mnihhgk14,bamk14,liylzwx17,samrudhdhicj22} is essential when all available information is expensive or unrealistic to process, e.g., scene classification for high-resolution satellite images~\citep{0009LCL19, samrudhdhicj22}. Fig.~\ref{fig:task-vis-digit-recognition} shows the task of hard attention for digit recognition on the Cluttered MNIST dataset \cite{mnihhgk14}.
    
    \item \textbf{Intrinsically motivated RL}. In this setting, an agent is trained using dense intrinsic rewards to explore the environment based on its curiosity about encountered states~\citep{pathak2017curiosity} or to discover diverse skills based on their novelty~\citep{Gregor17VariationalIntrinsic, Eysenbach2019DiversityAllYou, Strouse2022LearningMoreSkills}. Fig.~\ref{fig:task-vis-skill-discovery} shows the task of unsupervised skill discovery in a four-room environment \cite{Strouse2022LearningMoreSkills}.
    
    \item \textbf{Task-oriented active vision}. This is an emerging research topic with the goal of endowing embodied agents with high-level perception and reasoning capabilities. The agent actively changes its egocentric view to collect information for achieving downstream tasks, e.g., question answering~\citep{GordonKRRFF18,DengGGZ0021,liwklz0w21}, object recognition~\citep{YangRXCCPB19}, or scene description~\citep{tan0gz020}. Fig.~\ref{fig:task-vis-object-counting} shows the task of robotic object counting in occlusion scenarios.
\end{asparaenum}

\subsection{Optimization}
In IRRL, the policy and the discriminator are optimized simultaneously with different optimization objectives.

\subsubsection{Policy Optimization}
The optimization objective of policy learning in IRRL can be formulated from two perspectives, which are accuracy maximization and mutual information maximization.

\paragraph{Accuracy maximization.}
This is an intuitive formulation, where the policy of the agent is optimized to maximize the expectation of an accuracy-based reward
\begin{equation} \label{eq:accuracy_based_reward} r_{\text{acc}} = \mathds{1}_{y} \left[ \argmax_{y^\prime \in \mathcal{Y}} q_{\phi}(y^\prime\mid\tau) \right],
\end{equation}
where $\mathcal{Y}$ is a set of possible labels and $\mathds{1}_{y}[x]$ is an indicator function that returns 1 if $x$ is the target label $y$, 0 otherwise. This formulation has been widely used in existing works on hard attention~\citep{mnihhgk14,KingmaB14,liylzwx17}, embodied agents~\citep{GordonKRRFF18,YangRXCCPB19}, and robotics~\citep{lakomkinzwmw18,liwklz0w21}. However, an obvious disadvantage of the accuracy-based reward is that it cannot faithfully reflect the discriminator's uncertainty about the observations collected by the reinforcement learner, which makes learning slow and leads to suboptimal performance (cf.~Sec.~\ref{sec:experiments}). Therefore, it will be analyzed only empirically in this paper.

\paragraph{Mutual information maximization.}
Mutual information is commonly used to estimate the relationship between pairs of random variables. 
The objective of mutual information maximization has been utilized in the realm of unsupervised skill discovery~\citep{Gregor17VariationalIntrinsic, Eysenbach2019DiversityAllYou, Strouse2022LearningMoreSkills}. We generalize it to the optimization objective of IRRL.

Given a target label $y$ and a trajectory $\tau$ sampled from $p(y)$ and $\pi_\theta$ respectively, their mutual dependency can be obtained by the KL-divergence of their joint distribution $p(y, \tau)$ and the product of their marginal distributions $p(y)p(\tau)$:
\begin{align}\label{eq:mutual_info}
  I(y;\tau) &:= D_\mathbb{KL}(p(y, \tau)\mathrel{\|} p(y)p(\tau)) \\
            &= \mathbb{E}_{\tau \sim \pi_{\theta}, y \sim p(y)} \left [\log p(y\mid\tau) - \log p(y) \right],\nonumber
\end{align}
which is also known as Shannon's mutual information between $y$ and $\tau$ and which reaches its maximum if the full knowledge of $y$ can be deduced from $\tau$. 
In this equation, $p(y\mid\tau)$ is the oracle posterior probability that reflects the information sufficiency of observations for discrimination. 
It can be interpreted as being generated by an \emph{oracle discriminator}, a conceptual term utilized for the theoretical formulation. 
If $p(y\mid\tau)$ is known, then by defining 
\begin{equation*}
r_{\log}^{*}:=\log p(y\mid\tau) - \log p(y) \label{eq:oracle-reward}
\end{equation*}
as the reward for an RL algorithm involving $\pi_\theta$, one can maximize $I(y;\tau)$, i.e., $\pi_{\theta}$ generates trajectories for an optimal discrimination of the target label $y$.

Because the oracle discriminator $p(y \mid \tau)$ is not available in practice, we can replace $p(y \mid \tau)$ with a neural network $q_\phi(y \mid \tau)$ with trainable parameters $\phi$ and define the reward as
\begin{equation}
    \label{eq:log_reward}
    r_{\log} = \log q_\phi (y \mid \tau) - \log p(y),
\end{equation}
and maximize the Barber-Agakov variational lower bound of $I(y; \tau)$~\citep{Barber:2003aa}:
\begin{equation*}
    \label{eq:surrogate_mutal_info}
    I_{\text{BA}}(y; \tau) := \mathbb{E}_{\tau \sim \pi_{\theta}, y \sim p(y)} [\log q_\phi (y\mid\tau) - \log p(y)].
\end{equation*}

\subsubsection{Discriminator Optimization}
\label{sec:dicr-optim}

Concurrent with policy learning, the discriminator $q_{\phi}(y \mid \tau)$ is trained to better approximate $p(y \mid \tau)$. To this end, instead of the cross-entropy loss
\begin{equation*}
    - \mathbb{E}_{\tau \sim \pi_{\theta}, y \sim p(y)} \left[ p(y \mid \tau) \log q_{\phi}(y \mid \tau) \right],
\end{equation*}
which involves the oracle discriminator $p(y \mid \tau)$, a proxy cross-entropy loss $- \mathbb{E}_{\tau \sim \pi_{\theta}, y \sim p(y)} \log q_{\phi}(y \mid \tau)$
is used in practice, which is equivalent to assuming $p(y \, | \, \tau)=1$, i.e., assuming that $\tau$ contains sufficient information for deducing $y$ with the oracle discriminator.

\subsection{The Issue of Reward Noise}
\label{sec:noisy-reward}

As the trainable discriminator $q_{\phi}(y \mid \tau)$ only approximates the oracle discriminator $p(y \mid \tau)$, it inevitably introduces noise $\epsilon_{\log}$ in the reward $r_{\log}$ in Eq.~\eqref{eq:log_reward}, which is given by
\begin{equation*}
    \epsilon_{\log} = r_{\log} - r_{\log}^\star = \log q_\phi (y\mid\tau) - \log p(y\mid\tau).
\end{equation*}

\begin{figure}[ht]
    \centering
    \includegraphics[width=\linewidth]{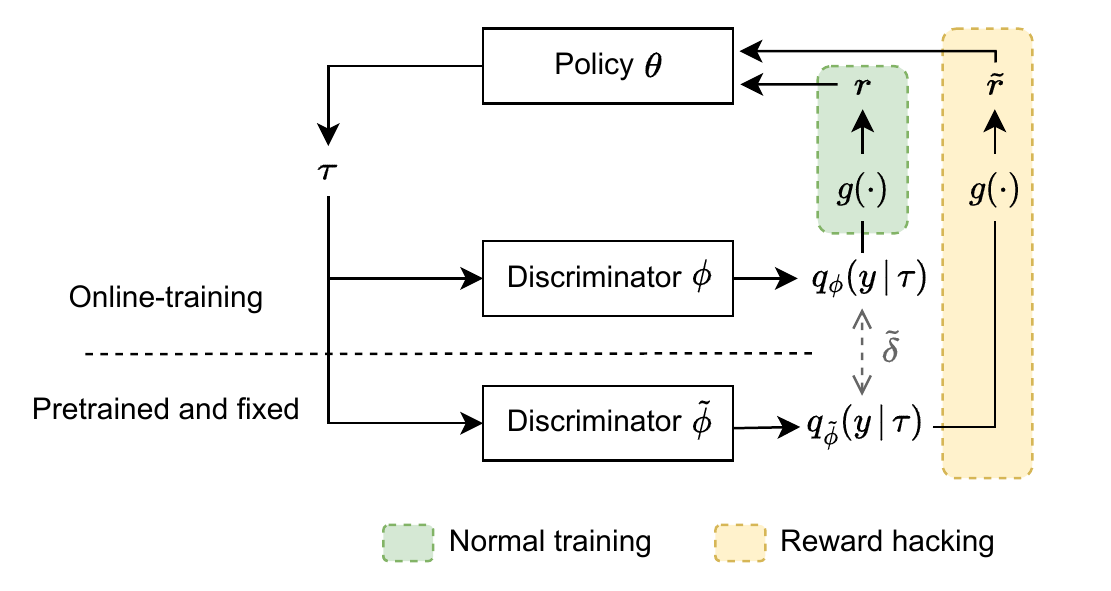}
    \caption{\ml{Illustration of the experimental setup of reward hacking. 
    In normal training, the reward is produced based on the posterior probability estimated by an online-training discriminator $\phi$. In training with reward hacking, the reward stems from a pretrained and fixed discriminator $\tilde{\phi}$. $\tilde{\delta}$ indicates the difference between a pair of posterior probabilities estimated by $\phi$ and $\tilde{\phi}$. }
    }
    \label{fig:diagram-reward-hacking}
\end{figure}

To demonstrate the negative impact of reward noise on the learning process (cf.~Fig.~\ref{fig:unstable-training-loop}), we conduct \emph{reward hacking} experiments, where we replace the trainable discriminator $q_\phi (y\mid\tau)$ with a pretrained one $q_{\tilde{\phi}} (y\mid\tau)$ that is obtained from a converged model to mimic the oracle discriminator $p(y \mid \tau)$. 
The setup of the reward hacking experiment is illustrated in Fig.~\ref{fig:diagram-reward-hacking}. 
We choose the digit recognition task as the target task (cf.~Fig.~\ref{fig:task-vis-digit-recognition}) and use the recurrent attention model (RAM)~\citep{mnihhgk14} (detailed information about this task and the model is given in Sec.~\ref{sec:experimental_setup} and Appendix~\ref{app:implementation}). 

\begin{figure}
    \centering%
    \includegraphics[width=\plotscale \linewidth]{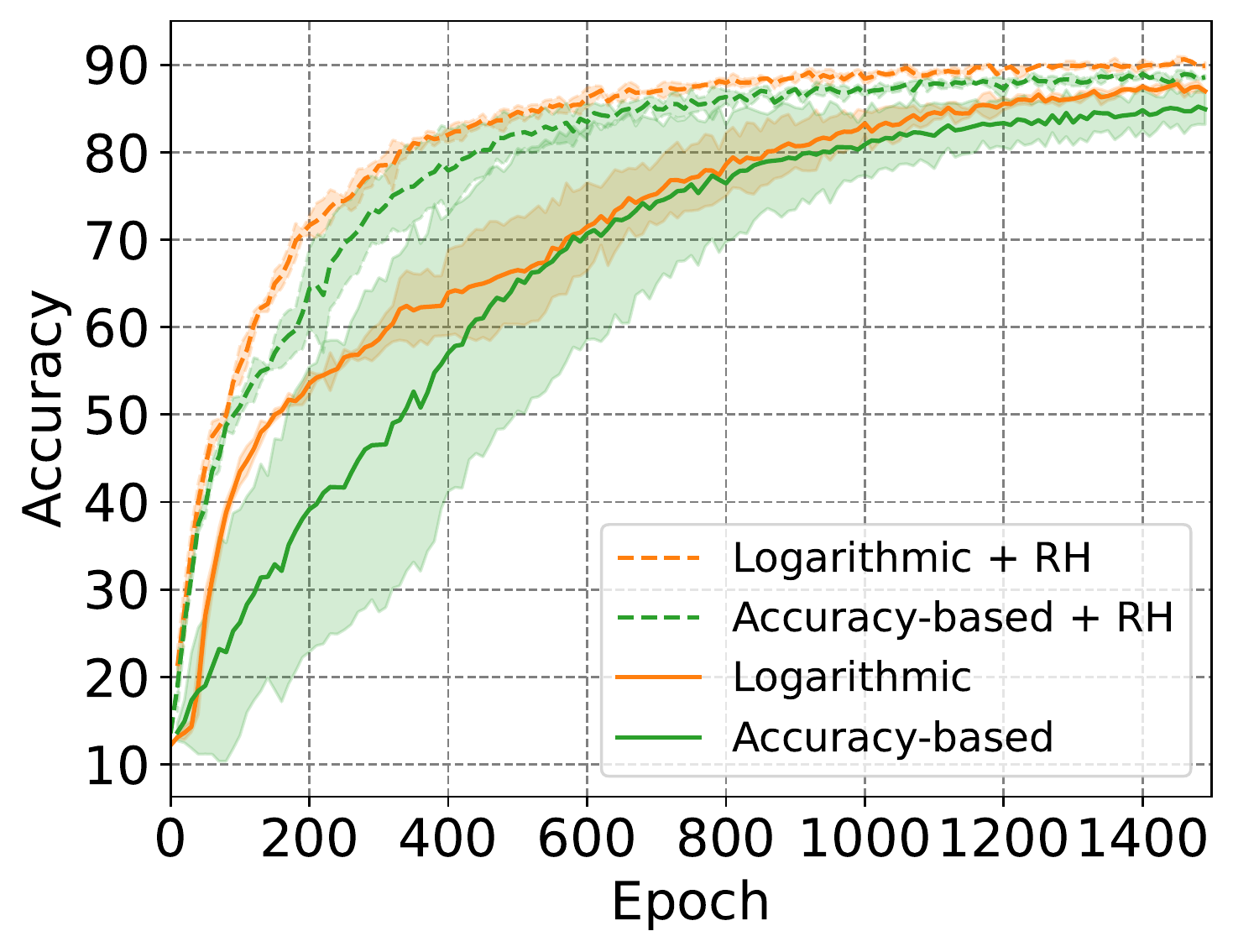}
    \caption{RAM trained using the accuracy-based and the logarithmic reward with and without reward hacking (RH). A model without reward hacking is subject to more noisy rewards and suffers from an unstable learning process, resulting in slower convergence and lower accuracy.}
    \label{fig:digit-recognition-reward-hacking-ram-acc-and-log}
\end{figure}

Fig.~\ref{fig:digit-recognition-reward-hacking-ram-acc-and-log} shows a plot of training curves when using different reward functions with and without reward hacking. 
\ml{As shown by the gap between the training curves when using an identical reward function with and without reward hacking}, the noise of an under-optimized discriminator influences the training process negatively. In this paper, we aim to narrow the gap by devising an effective reward. As we will see in the next section, a well-designed reward is key to stabilizing the learning process.

\section{Reward Noise Moderation}
\label{sec:noise-moderation}

In this section, we first analyze the reduction of the bias of the estimated reward and the variance of the reward noise and then propose a reward that alleviates the negative effect of reward noise and stabilizes the training process.

\subsection{Generalized Reward}

Since the noisy reward in Eq.~\eqref{eq:log_reward} is a transformation of the posterior probability $q_{\phi}(y \mid \tau)$, it is reasonable to study the effect of a series of transformations of $q_{\phi}(y \mid \tau)$ as long as they agree on the same optimal objective. Based on the logarithmic transformation in Eq.~\eqref{eq:log_reward}, the \emph{generalized reward} is defined as
\begin{equation} 
\label{eq:r-g-function}
    r_g = g\left[q_\phi(y \mid \tau)\right] - g\left[p(y)\right]
\end{equation}
and the \emph{generalized oracle reward} as 
\begin{equation*} 
r_g^{*} = g\left[p(y \mid \tau)\right] - g\left[p(y)\right],
\end{equation*}
where $g$ is an increasing function (e.g., $\log$), such that maximizing $g(\cdot)$ leads to the maximization of the mutual information $I(y;\tau)$.
When selecting the appropriate function, it is important to consider both its ability to transmit information and its ability to moderate noise. The former ensures that the maximization of mutual information can be achieved efficiently, while the latter helps to reduce the impact of reward noise.\footnote{The transformation $g(\cdot)$ can also be motivated by the $f$-mutual information objectives (see Appendix~\ref{sec:proof_linear_reward_function_chi_square}).}

\subsection{Generalized Reward Noise}
To analyze the noise in the generalized reward $r_g$ we apply the second-order Taylor approximation to the \emph{generalized reward noise} 
\begin{equation*}
\epsilon_{g} := r_{g} - r_{g}^{*} = g\left[q_\phi(y \mid \tau)\right] - g\left[p(y \mid \tau)\right]    
\end{equation*}
at point $p(y \mid \tau)$.
By defining $\delta := q_\phi (y \mid \tau) - p(y \mid \tau)$ as the \emph{discriminator noise}, we have as the expectation of the reward noise (equivalently, the bias of the reward estimator)
\begin{equation} \label{eq:noise_bias_taylor}
    \begin{split}
      \mathbb{E}[\epsilon_{g}]
      \approx \: & g'(p(y \mid \tau)) \mathbb{E}[\delta] + \frac{1}{2!} g''(p(y \mid \tau)) \mathbb{E}\left[\delta^2\right],
    \end{split}
\end{equation}
and as the variance of the reward noise
\begin{align} \label{eq:noise_variance_taylor}
      \mathbb{V}[\epsilon_{g}] &\approx (g'(p(y \mid \tau)))^2 \mathbb{V}[\delta] + (\frac{1}{2!}  g''(p(y \mid \tau)))^2 \mathbb{V}\left[\delta^2\right] \nonumber\\
                 & \quad+ g'(p(y \mid \tau)) g''(p(y \mid \tau)) \mathrm{Cov}\left[\delta, \delta^2\right].
\end{align}
Our goal is to mitigate the impact of the reward noise by minimizing the expectation and the variance of the noise. 
This is expected to be achieved especially at the early learning stage when the issue of the unstable training loop is severe because both the discriminator and the policy are immature: the trajectory collected by the policy contains little information for discrimination, and the estimated posterior of the discriminator cannot reflect the sufficiency of information collected by the policy.
To this end, in the following sub-sections, we theoretically and empirically analyze Eq.~\eqref{eq:noise_bias_taylor} and Eq.~\eqref{eq:noise_variance_taylor} and investigate reward functions.

\begin{figure}[ht]
  \centering%
  \includegraphics[scale=0.42]{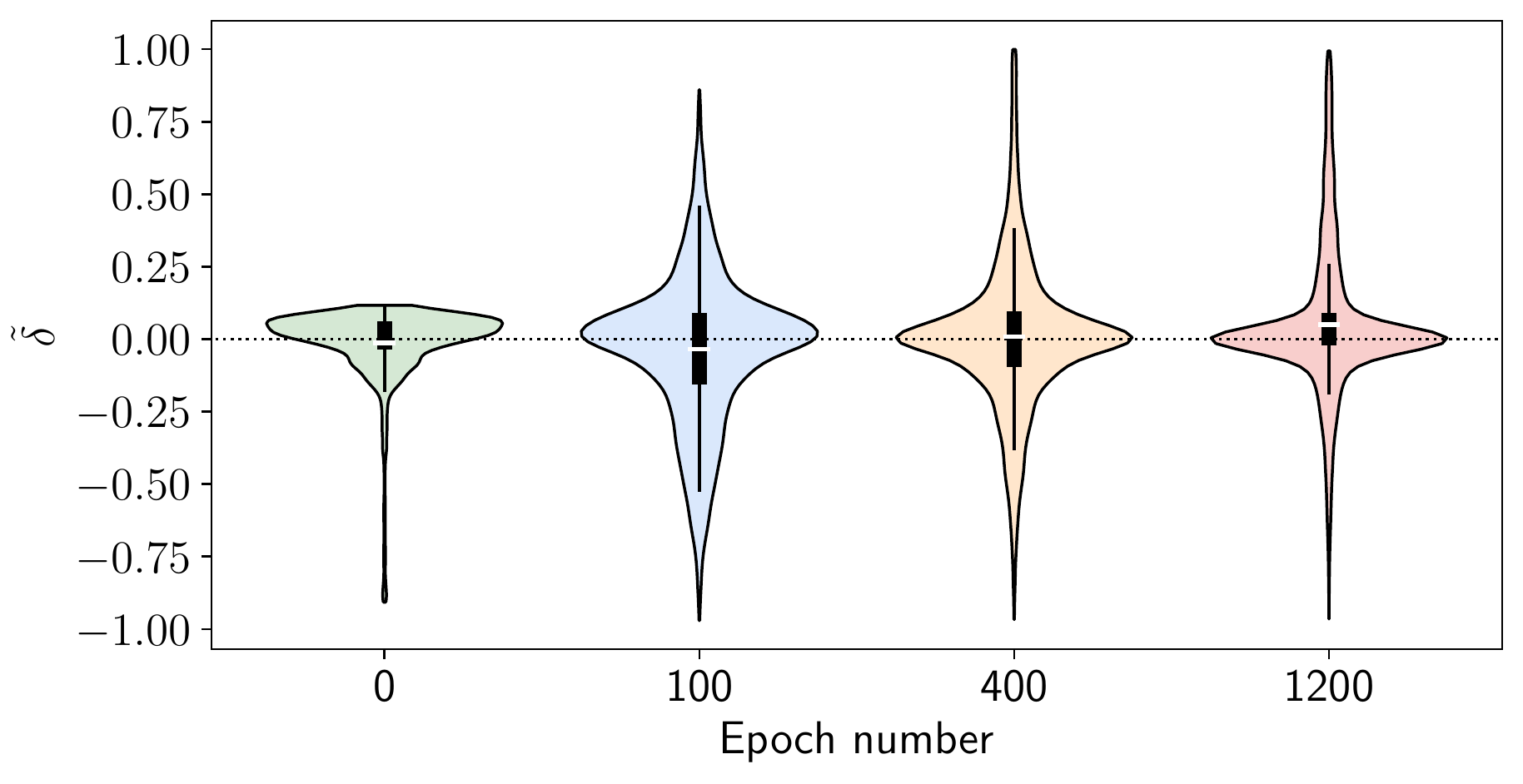}
  \caption{\ml{Violin plots of the approximated discriminator noise $\tilde{\delta}$ in the training process of RAM, of which the policy is trained using the logarithmic reward function (cf. Eq.~\eqref{eq:log_reward}). The small white bar indicates the mean of the noise. The thick vertical line represents the interquartile range and the thin vertical line represents the area between the upper and lower adjacent values. }} 
  \label{fig:noisy-visualization-log}
\end{figure}

\ml{
\subsection{Characterization of the Discriminator Noise} \label{sec:noise-characterization}
We make hypotheses regarding the distribution characteristics of $\delta$, which is necessary to analyze the expectation and variance of the reward noise $\epsilon_g$ according to Eq.~(\ref{eq:noise_bias_taylor}) and Eq.~(\ref{eq:noise_variance_taylor}).
We hypothesize that the expectation of the discriminator noise $\delta$ is zero, i.e., $\mathbb{E}[\delta] = 0$, and the distribution of $\delta$ is symmetric. 

We conduct an empirical study of the distribution of the discriminator noise following the setup of the reward hacking experiment (cf.~Fig.~\ref{fig:diagram-reward-hacking}).
Instead of using a pretrained discriminator to interfere in the training process, 
we visualize the approximated discriminator noise $\tilde{\delta}$ during normal training. $\tilde{\delta}$ is the difference between the posterior probabilities estimated by the online-training discriminator and the pretrained discriminator, i.e., $\tilde{\delta} =  q_\phi(y \mid \tau) - q_{\tilde{\phi}}(y\mid\tau) \approx \delta$.

Fig.~\ref{fig:noisy-visualization-log} demonstrates violin plots of the discriminator noise at four training epochs (the model converges at about 1200 epochs). 
Each violin plot is drawn from 1000 random samples from the testing dataset.
We can observe that the mean of the noise is close to zero at different training stages, i.e., $\mathbb{E}[\delta] \approx 0$, and the plots are almost symmetrical with respect to the average noise except at the very beginning when the model weights are being updated after random initialization.

We assume that the noise characteristics are generalized to other problems of IRRL because of the shared high-level abstraction among them (cf. Fig.~\ref{fig:elaborated-agent-environement-interaction-with-sl-and-noisy-rewards}).
Characteristics of the discriminator noise when using other reward functions are given in Appendix \ref{sec:discriminator-noise-visulization}. 
}


\subsection{Linear Reward} \label{sec:linear-reward-function}

Considering the impact of $g(\cdot)$ on the Taylor approximation to $\mathbb{E}[\epsilon_g]$ and $\mathbb{V}[\epsilon_g]$ in Eq.~(\ref{eq:noise_bias_taylor}) and Eq.~(\ref{eq:noise_variance_taylor}), we propose a linear reward
\begin{equation} \label{eq:linear_reward} r_\text{lin} = q_\phi (y \mid \tau) - p(y),
\end{equation}
instead of the commonly applied logarithmic reward $r_{\log}$, to stabilize IRRL. The corresponding expectation and variance of the noise are $\mathbb{E}[\epsilon_\text{lin}] = \mathbb{E}[\delta] = 0$ and $\mathbb{V}[\epsilon_\text{lin}] = \mathbb{V}[\delta]$, respectively.
The linear reward enjoys lower reward bias than the logarithmic reward, since 
\begin{equation*}
|\mathbb{E}[\epsilon_{\log}]| \approx  \frac{1}{2! \, p^2(y \mid \tau)} \mathbb{E}[\delta^2] > 0 = |\mathbb{E}[\epsilon_{\text{lin}}]|.
\end{equation*}
Furthermore, the variance of $r_{\text{lin}}$ is low and stable compared with the variance of logarithmic reward $r_{\log}$, which suffers from high variance 
\begin{equation*}
\mathbb{V}[\epsilon_\text{log}] \approx p^{-2}(y \mid \tau) \mathbb{V}[\delta] + (\frac{1}{2! \, p^2(y \mid \tau)})^2 \mathbb{V}[\delta^2],
\end{equation*}
since $p(y \mid \tau) < 1$ in most cases and is dependent on the training policy. (A detailed derivation is given in Appendix~\ref{app:proof_comparison}. The evaluation of various $g$ functions is given in Appendix~\ref{sec:evaluation-g-functions}.)

\subsection{Clipped Linear Reward} \label{sec:clipped-linear-reward-function}
The issue of reward noise is not fully tackled by using the linear reward. Given a target label $y$, it is intuitive to assume that the posterior probability $p(y \mid \tau)$ of an oracle discriminator should be, in most cases, equal or larger than the prior $p(y)$, as $y$ is a cause of the trajectory $\tau$. However, a discriminator $q_\phi$ may return a posterior probability $q_\phi(y \mid \tau)$  lower than $p(y)$, especially at the early training stage when both the policy and the discriminator are under-optimized. 


Since we expect $q_\phi (y \mid \tau)$ to be close to $p(y \mid \tau)$, we replace the term $q_\phi(y \mid \tau)$ of $r_{\text{lin}}$ in Eq.~\eqref{eq:linear_reward} with $\max (q_\phi(y \mid \tau), p(y))$ to integrate the prior knowledge and define the \textit{clipped linear reward} as
\begin{align}
\label{eq:linear_reward_clipped}
   \overline{r_{\text{lin}}} &:= \max (q_\phi(y \mid \tau), p(y)) - p(y)\nonumber\\
   & \:=\max (q_\phi (y\mid\tau) - p(y), 0).
\end{align}
Similar clipping techniques are empirically found to be beneficial when applied to the logarithmic reward~\cite{Strouse2022LearningMoreSkills}. In this paper, we go further with an analysis of reward functions from the perspective of noise moderation and achieve better performance with the proposed reward. 
The proposed clipped linear reward has a similar shape to the rectified linear unit (ReLU) activation function~\citep{Nair2010RectifiedLinearUnits} which preserves information about relative intensities in multiple layers of deep neural networks. Likewise, the clipped linear reward function can robustly preserve information that travels from an internal discriminator to the policy network.

\section{Experiments} \label{sec:experiments}
In this section, we conduct experiments to evaluate the effectiveness of the proposed method on the three aforementioned tasks in Sec.~\ref{sec:related_work_internally_rewarded_rl}.
We first introduce experimental setups and baselines.
Then, we compare the proposed clipped linear function with multiple baselines including state-of-the-art methods. 
Finally, we conduct reward hacking experiments on the clipped linear reward function to visualize its capability in reducing the impact of reward noise.

\begin{figure}[ht]
  \centering%
  \begin{subfigure}[ht]{\plotscale \linewidth}
    \includegraphics[width=\textwidth]{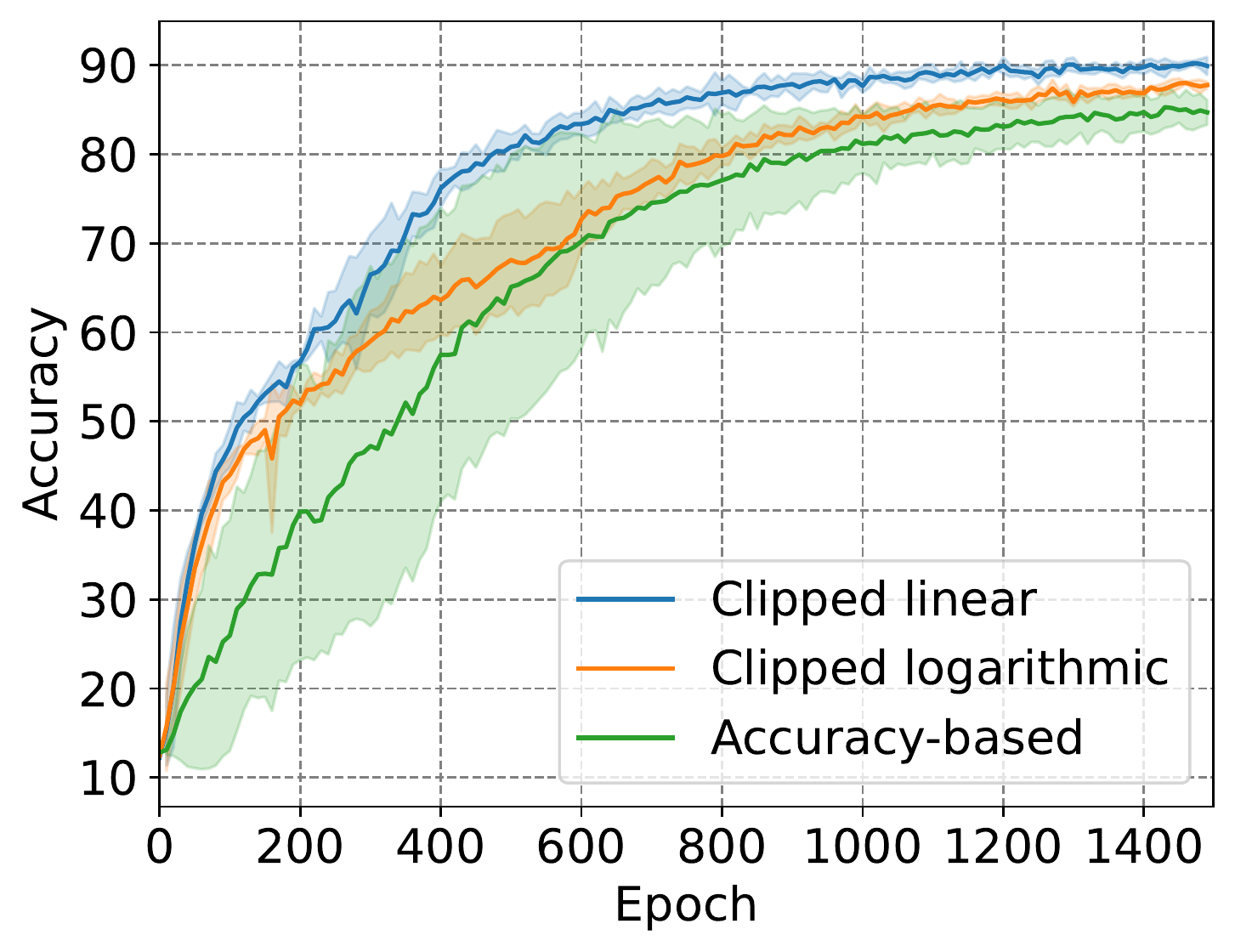}
    \caption{RAM}
  \end{subfigure}
  \begin{subfigure}[ht]{\plotscale \linewidth}
    \includegraphics[width=\textwidth]{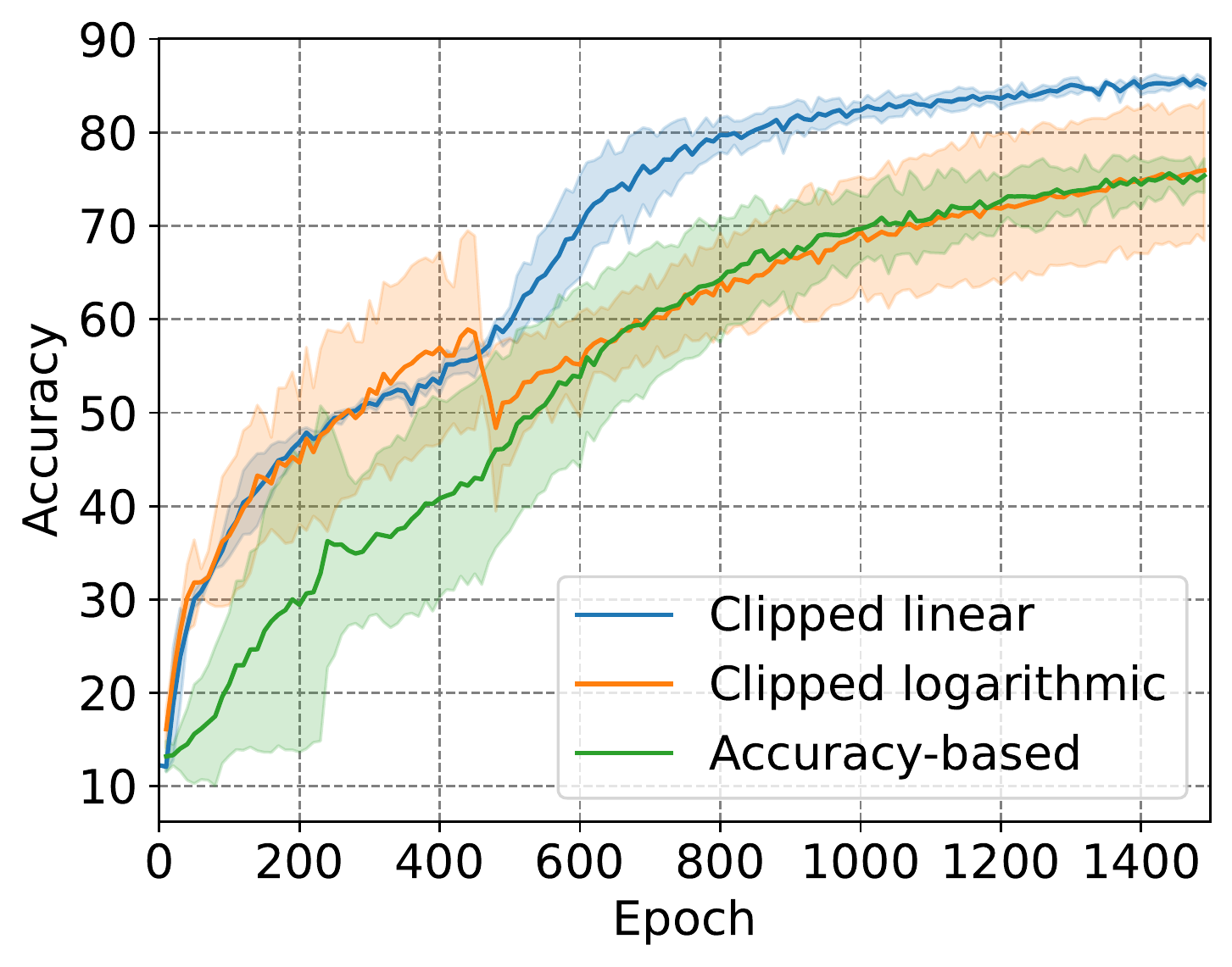}
    \caption{DT-RAM}
  \end{subfigure}
  \caption{Comparison between the clipped linear reward function (\colorrule[myblue]{0.2cm}{0.2cm}) with baselines, including the clipped logarithmic (\colorrule[myorange]{0.2cm}{0.2cm}) and the accuracy-based (\colorrule[mygreen]{0.2cm}{0.2cm}) reward function, on the task of hard attention for digit recognition using RAM and DT-RAM. All the experiments in this paper ran over three random seeds. Lines and shaded areas show the mean and standard deviation over multiple runs. }
  \label{fig:digit-recognition-baseline-comparison}
\end{figure}

\subsection{Experimental Setup} \label{sec:experimental_setup}

\paragraph{Hard attention for digit recognition.} 

We adopt the dataset configuration of~\citeauthor{mnihhgk14} (\citeyear{mnihhgk14}),
and use two basic models for this task: the recurrent attention model (RAM)~\citep{mnihhgk14} and the dynamic-time recurrent attention model (DT-RAM)~\citep{liylzwx17}.
RAM performs a fixed number of movement steps before performing the final digit recognition, while the policy of DT-RAM learns to terminate the exploration before reaching a maximum number of movement steps.
The performance of the agent is evaluated using the accuracy of the digit recognition.

\paragraph{Unsupervised skill discovery.}
We use the same experimental setup and basic model on the four-room environment as in the work of the discriminator disagreement intrinsic reward (DISDAIN)~\citep{Strouse2022LearningMoreSkills}.
The performance of the agent is evaluated using the number of learned skills.

\paragraph{Robotic object counting.}
The setup is based on the task of object existence prediction~\citep{liwklz0w21}.
We use their model and train it using PPO~\citep{schulmanwdrk17} instead of REINFORCE for higher efficiency.
The performance of the agent is evaluated using the accuracy of object counting.

Details of the environments and model implementations can be found in Appendix \ref{app:environments} and \ref{app:implementation}.

\subsection{Baselines} \label{sec:alter-methods}
We compare the proposed clipped linear reward function with alternative reward functions.
The first is the \emph{accuracy-based reward function} $r_\text{acc}$ in Eq.~\eqref{eq:accuracy_based_reward}.
The second is the \emph{logarithmic reward function} based on Shannon's mutual information.
Instead of using the original logarithmic reward function (Eq.~\eqref{eq:log_reward}), we use a clipped variant, i.e.,
$\overline{r_{\log}} = \max (\log q_\phi (y \mid \tau) - \log p(y), 0)$, for fair comparison with our clipped linear reward function. We found that reward clipping generally results in similar or better performance in our experiments, which is consistent with the empirical finding by~\citeauthor{Strouse2022LearningMoreSkills} (\citeyear{Strouse2022LearningMoreSkills}).
The empirical study of reward clipping is provided in Appendix~\ref{sec:reward-clipping}.
On the skill discovery task, we additionally compare our reward function with the state-of-the-art \emph{DISDAIN reward function} ~\citep{Strouse2022LearningMoreSkills} (see Appendix \ref{app:disdain} for details), which was designed specifically to mitigate the pessimistic exploration issue in this task.
\subsection{Results}
\begin{figure}[ht]
  \centering%
  \begin{subfigure}[ht]{\plotscale \linewidth}
    \includegraphics[width=\textwidth]{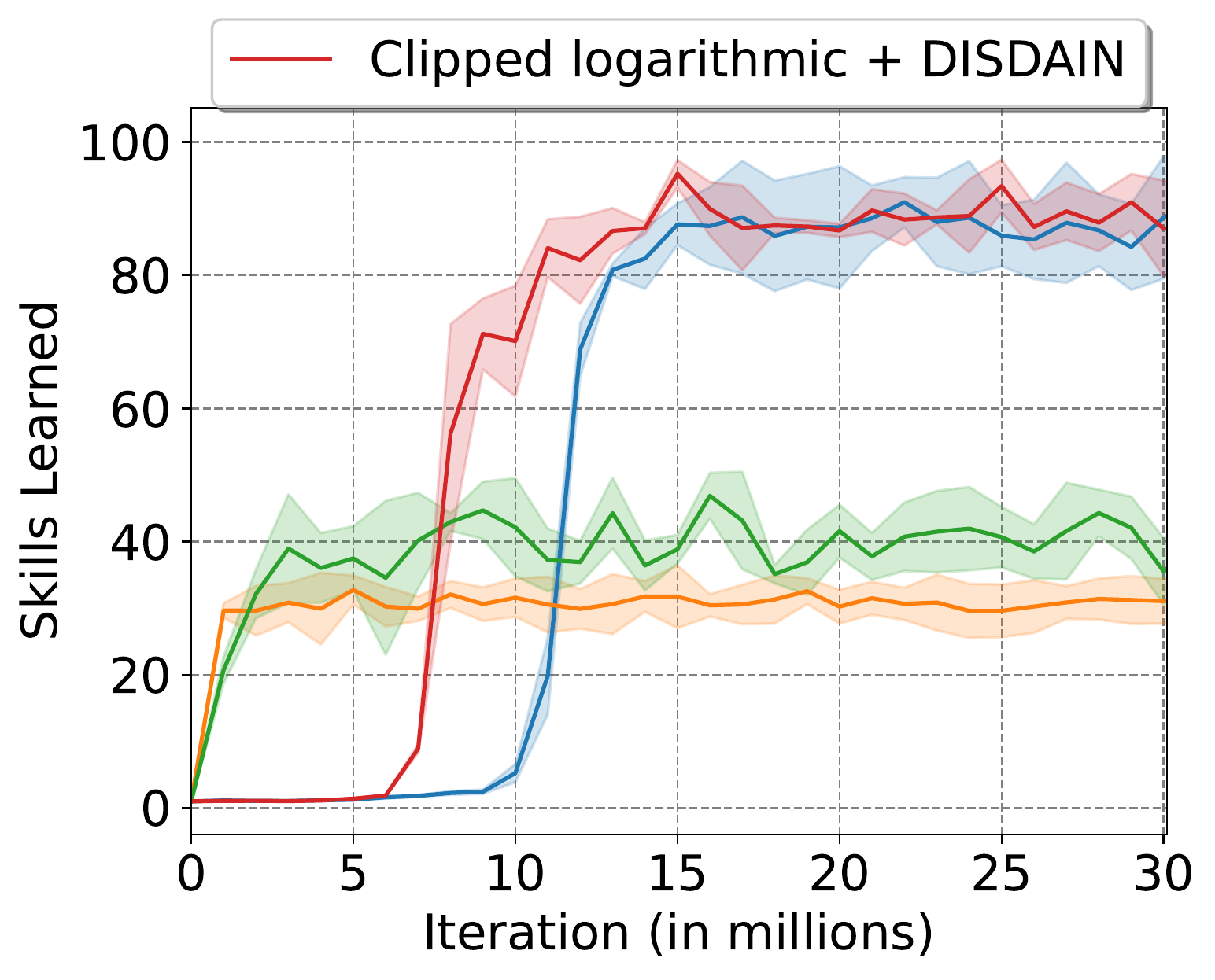}
    \caption{Unsupervised skill discovery}
    \label{fig:skill-discovery-baseline-comparison}
  \end{subfigure}
  \hfill
  \begin{subfigure}[ht]{\plotscale \linewidth}
    \includegraphics[width=\textwidth]{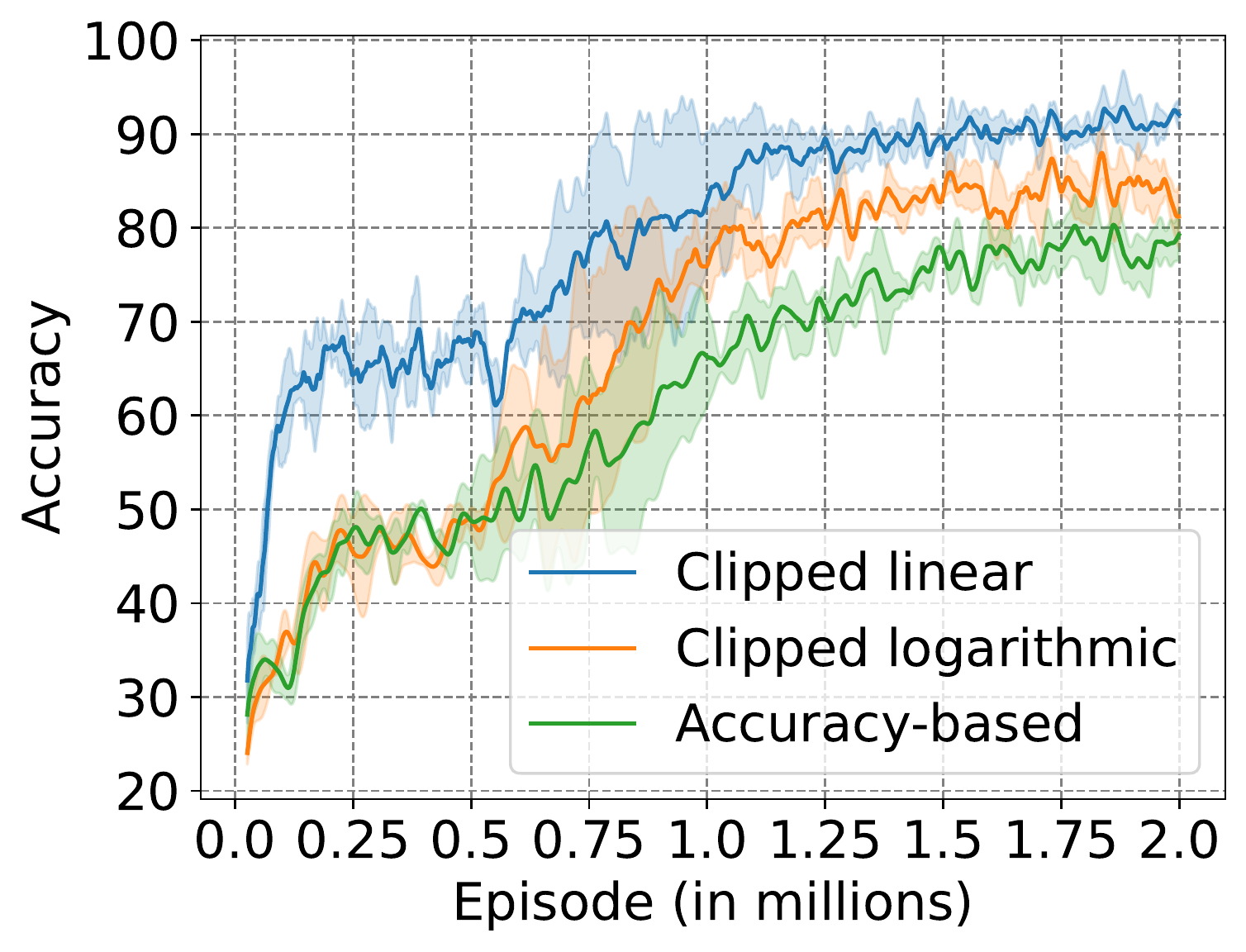} 
    \caption{Robotic object counting}
    \label{fig:object-counting-baseline-comparison}
  \end{subfigure}
  \caption{Comparison with baselines on the tasks of unsupervised skill discovery and robotic object counting. On the unsupervised skill discovery task, the auxiliary DISDAIN reward (\colorrule[myred]{0.2cm}{0.2cm}) is compared. Legends are shared between the two sub-figures. 
  }
  \label{fig:skill-discovery-and-object-counting-baseline-comparison}
\end{figure}
Fig.~\ref{fig:digit-recognition-baseline-comparison} shows that both RAM and DT-RAM trained using the clipped linear reward function achieve the highest accuracy and fastest training speed.
Furthermore, the small blue shaded areas indicate that multiple runs using the clipped linear reward function are consistent with each other, which suggests high stability of the training process. 

Fig.~\ref{fig:skill-discovery-baseline-comparison} demonstrates that the clipped linear reward function outperforms both the clipped logarithmic reward function and the accuracy-based reward function by a large margin and achieves almost the same performance as DISDAIN.
We note that the DISDAIN method depends on an ensemble of discriminators and needs more hyper-parameters to tune, e.g., the weight of the DISDAIN reward and the number of ensemble members, while our method is much simpler.
Fig.~\ref{fig:object-counting-baseline-comparison} shows that the clipped linear reward function also benefits the challenging robotic object counting task by making the model converge faster and achieve the highest final accuracy. 

We can see that the clipped linear reward function generally outperforms the logarithmic and the accuracy-based reward function. 
The improvement is significant on the skill discovery task, which makes sense according to our theoretical analysis in Sec.~\ref{sec:linear-reward-function}.
Since the number of possible discrimination classes in the skill discovery task (128 classes) is much larger than that of other tasks (10 classes in the digit recognition task, and 7 classes in the robotic object counting task), $p(y \mid \tau)$ tends to be closer to zero when trajectory $\tau$ contains a small amount of information for discrimination in the skill discovery task.
Thus the expectation and variance of the noise of the logarithmic reward function are larger, resulting in a severer unstable training issue, while our clipped linear reward function resulting in low expectation and variance of the reward noise still performs well. 

Interesting case studies for the digit recognition and object counting tasks are given in Appendix \ref{app:case-study}.
An intuitive comparison of state occupancy in the unsupervised skill discovery task is given in Appendix \ref{app:state-occupancy}.

\subsection{Effect of Noise Moderation}
\begin{figure}[t]
  \centering%
  \includegraphics[width=\plotscale \linewidth]{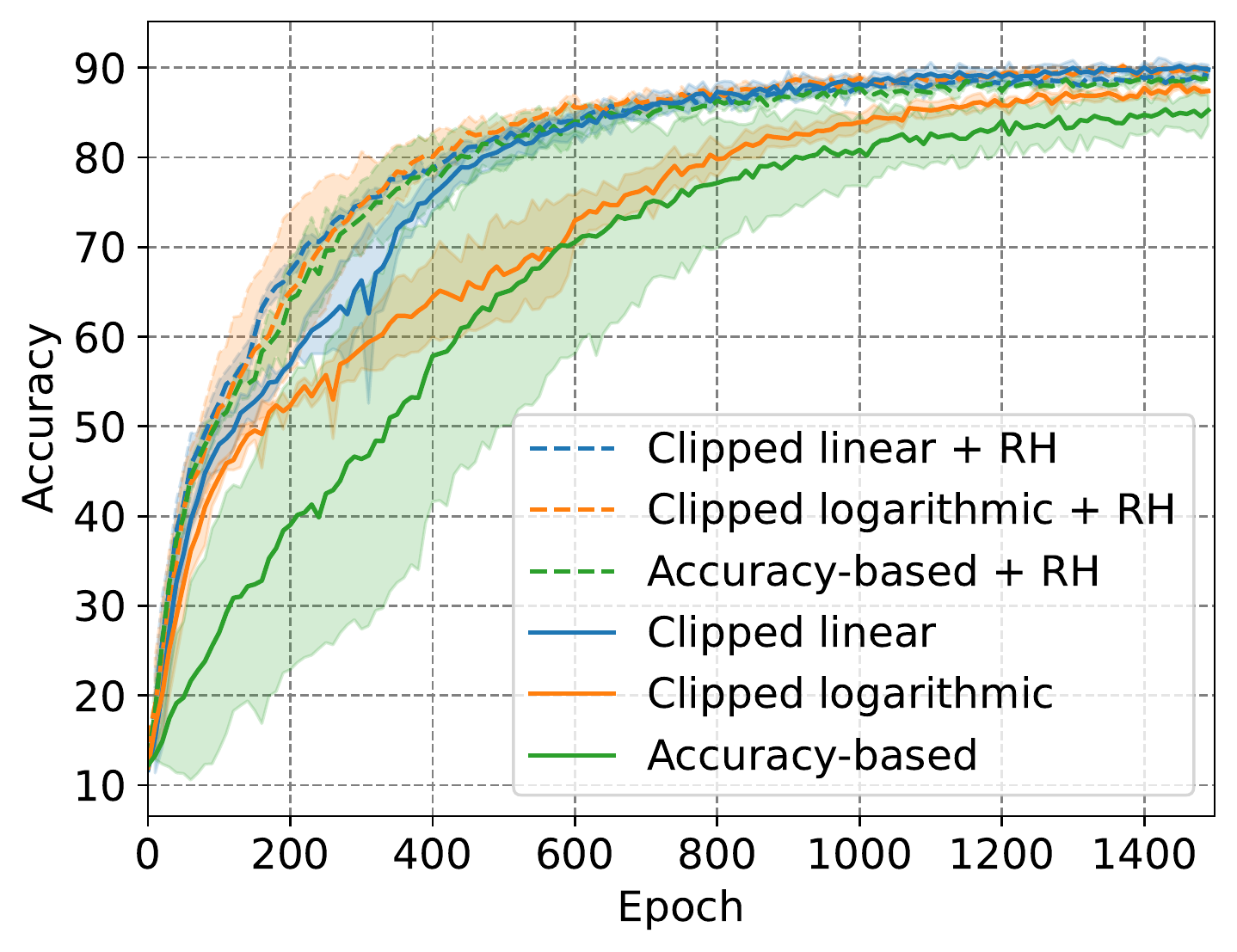}
  \caption{
  RAM trained using the three kinds of reward functions with and without reward hacking (RH). 
  The clipped linear reward function achieves a much smaller gap between the training processes with and without reward hacking.
  }
  \label{fig:digit-recognition-reward-hacking-ram}
\end{figure}

Following the experimental setup in Sec.~\ref{sec:noisy-reward}, we conduct reward hacking experiments using the clipped linear reward to visualize its capability in narrowing the gap between training processes with and without reward hacking (cf.~Fig.~\ref{fig:diagram-reward-hacking}). 
Fig.~\ref{fig:digit-recognition-reward-hacking-ram} shows the training curves. In order to facilitate a comprehensive comparison, we incorporate training curves when using the accuracy-based and the logarithmic reward (cf.~Fig.~\ref{fig:digit-recognition-reward-hacking-ram-acc-and-log}) into the figure. 
We can see from Fig.~\ref{fig:digit-recognition-reward-hacking-ram} that when using reward hacking, all three rewards perform similarly (see dashed lines).
This suggests that the linear function performs as well as the logarithmic function in terms of information transmission.
However, when not using reward hacking, the training curve of the clipped linear reward is much closer to the training curve of using reward hacking, compared to the other two rewards.
This suggests that the advantage of the clipped linear reward function is due to the reduction of the impact of reward noise.

\section{Discussion}
\subsection{Interpretation from the Information-theoretic Perspective}
The linear reward function has specific meanings
from an information-theoretic perspective. 
It can be derived from the optimization objective of maximizing the $\chi^2$-divergence, one of the $f$-mutual information measures~\citep{Csiszar72ClassMeasures,Esposito20RobustGeneralization}, instead of the commonly used KL-divergence corresponding to Shannon's mutual information~\citep{Shannon1948MathematicalTheoryCommunication} (cf. Eq.~\eqref{eq:mutual_info}). The derivation is provided in Appendix~\ref{sec:proof_linear_reward_function_chi_square}.

In recent years, $f$-mutual information has been studied in many deep learning applications, such as generative models~\citep{Nowozin16FGANTraining, JaimeUnifiedfVAEGAN2022}, representation learning~\citep{AliInfoMaxVAE2020, AbstreiterDiffusionRepresentationLearning2021}, image classification~\citep{WeifDivergence2021}, imitation learning~\citep{zhangfGAIL2020}, etc.~\citeauthor{WeifDivergence2021} (\citeyear{WeifDivergence2021}) suggested that a properly defined $f$-divergence measure is robust with label noise in a classification task, which is related to our finding 
that the $\chi^2$-mutual information is a more robust information measure against the inherent noise in the policy learning of  IRRL compared to Shannon's mutual information. This leads to interesting future work on investigating principles for selecting the optimal $f$-mutual information measure, and the possibility of using other $f$-mutual information measures for achieving more stable IRRL.

\subsection{Limitations and Future Work}
This work is an early step towards stabilizing IRRL. Some identified limitations potentially lead to interesting future work. 
First, we evaluate the efficiency of the suggested reward functions within a subset of IRRL scenarios. It is appealing to study the generaliablity and explore the potential adaptions across a wider spectrum of applications, e.g., RLHF for large language models finetuning.
Second, we only consider classification-based reward models but not regression-based ones.
A unified guideline for designing reward functions in both cases is significant.
Third, we stabilize the training process of IRRL from the perspective of reducing the impact of reward noise without explicitly considering reducing the impact of insufficient observations (see Fig.~\ref{fig:elaborated-agent-environement-interaction-with-sl-and-noisy-rewards}). 
An integrated method considering both issues should lead to a more optimal solution. 

\section{Conclusion}
In this work, we formulate a class of RL problems with internally rewarded RL where a policy and a discriminator functionally interact with each other and are simultaneously optimized. 
The inherent issues of noisy rewards and insufficient observations in the training process lead to an unstable training loop where neither the policy nor the discriminator can learn effectively. 
Based on theoretical analysis and empirical studies, we propose the clipped linear reward function to reduce the impact of reward noise. 
Extensive experimental results suggest that 
the proposed method can consistently stabilize the training process and achieve faster convergence and higher performance compared with baselines in diverse tasks.
Additionally, we give an interpretation of the use of the linear reward function from the information-theoretic perspective, which suggests interesting future work. 
As interest grows in integrating the capability of high-level prediction and low-level control of behaviors into a single model, for instance in embodied AI, robotics, and unsupervised RL, stable and efficient training of IRRL will be particularly relevant. 
We hope this work paves the way to achieving this goal. 

\section*{Acknowledgements}
We gratefully acknowledge support from the China Scholarship Council (CSC) and the German Research Foundation (DFG) under project Crossmodal Learning (TRR 169).

\newpage
\bibliography{mybib}
\bibliographystyle{icml2023}

\newpage
\appendix
\onecolumn

\section{Proofs}
\subsection{Noise of the Logarithmic Reward}\label{app:proof_comparison}
Based on the formulation of Eq.~\eqref{eq:noise_bias_taylor} and Eq.~\eqref{eq:noise_variance_taylor}, the expectation and variance of the reward noise when using the logarithmic reward ($\epsilon_\text{log}$) can be derived as follows: 

\begin{equation*}
\begin{split}
\mathbb{E}[\epsilon_\text{log}] = \: & g'(p(y \mid \tau)) \mathbb{E}[\delta] + \frac{1}{2!} g''(p(y \mid \tau)) \mathbb{E}[\delta^2] + \mathbb{E}[o(\delta^2)] \\
= \: & \frac{1}{p(y \mid \tau)}\mathbb{E}[\delta] - \frac{1}{2! \, p^2(y \mid \tau)} \mathbb{E}[\delta^2] + \mathbb{E}[o(\delta^2)]\\
\approx \: & - \frac{1}{2! \, p^2(y \mid \tau)} \mathbb{E}[\delta^2], \\
\\
\mathbb{V}[\epsilon_\text{log}] = \: & (g'(p(y \mid \tau)))^2 \mathbb{V}[\delta] + (\frac{1}{2!}  g''(p(y \mid \tau)))^2 \mathbb{V}[\delta^2] + g'(p(y \mid \tau)) g''(p(y \mid \tau)) \mathrm{Cov}[\delta, \delta^2] \\
& + \mathbb{V}[o(\delta^2)] + 2 g'(p(y \mid \tau)) \mathrm{Cov}[\delta, o(\delta^2)] + g''(p(y \mid \tau)) \mathrm{Cov}[\delta^2, o(\delta^2)] \\
\approx \: & (g'(p(y \mid \tau)))^2 \mathbb{V}[\delta] + (\frac{1}{2!}  g''(p(y \mid \tau)))^2 \mathbb{V}[\delta^2] + g'(p(y \mid \tau)) g''(p(y \mid \tau)) \mathrm{Cov}[\delta, \delta^2] \\
\approx \: & (g'(p(y \mid \tau)))^2 \mathbb{V}[\delta] + (\frac{1}{2!}  g''(p(y \mid \tau)))^2 \mathbb{V}[\delta^2] \\
= \: & \frac{1}{p^2(y \mid \tau)} \mathbb{V}[\delta] + (\frac{1}{2! \, p^2(y \mid \tau)})^2 \mathbb{V}[\delta^2]. \\
\end{split}
\end{equation*}

The variance is approximated using the fact that $\mathrm{Cov}[\delta, \delta^2] = \mathbb{E}[\delta^3] -\mathbb{E}[\delta]\mathbb{E}][\delta^2] = (\mu^3 + 3\mu \sigma^2 + \gamma \sigma^3) - \mu (\mu^2 + \sigma^2) =  2 \mu \sigma^2 + \gamma \sigma^3 \approx 0$,
where $\sigma^2 = \mathbb{E}[(\delta - \mu)^2]$ is the variance, and $\mu = \mathbb{E}[\delta]$ and $\gamma = \mathbb{E}[(\frac{\delta-\mu}{\sigma})^3]$ are the mean and skewness, which are both about zero due to the symmetry of the distribution of $\delta$ (Sec.~\ref{sec:noise-characterization}).

\subsection{Derivation of the Linear Reward Function by Maximizing the \texorpdfstring{$\chi^2$-divergence}{Lg}} \label{sec:proof_linear_reward_function_chi_square}

We use the optimization objective of maximizing the $f$-mutual information between the observation trajectory $\tau$ and the target class $y$ in place of the objective of Eq.~\eqref{eq:mutual_info} and obtain
\begin{equation*}
\label{eq:f-mutual-info}
    \begin{split}
      I_f(y;\tau) := \: & D_f({p(y,\tau)} \, || \, p(y)p(\tau)) \\
      = \: & \mathbb{E}_{\tau \sim \pi_{\theta}, y \sim p(y)} F \left( \frac{p(y \mid \tau)}{p(y)} \right),
    \end{split}
\end{equation*}
where $D_{f}(P \, || \, Q) := \mathbb{E}_{q(x)} f\left(\frac{p(x)}{q(x)} \right) = \mathbb{E}_{p(x)} F\left(\frac{p(x)}{q(x)}\right)$ is the $f$-divergence of two probability distributions $P$ and $Q$ on $X$, with $f: \mathbb{R}^{+} \rightarrow \mathbb{R}$ being a generic convex function satisfying $f(1)=0$, $F(x) :=  f(x)/x$ for simplicity of expectation over $P$ instead of $Q$ for later use, and $p(x)$ and $q(x)$ are probability density functions of $P$ and $Q$ respectively.
By choosing $f(x)=x \log x$, $f$-divergence becomes the well-known Kullback–Leibler divergence and, correspondingly, the $f$-mutual information is then Shannon's mutual information~\citep{Shannon1948MathematicalTheoryCommunication,Kinney14EquitabilityMutual,Belghazi18MutualInformation}.
Other typically used $f$-divergences and their expected mutual information over $p(x,y)$ are listed in Table \ref{table:div}. 
When using the $\chi^2$-divergence, i.e., $f(x) = (x-1)^2$, $f$-mutual information becomes
\begin{equation*} 
    \begin{split}
      I_f(y;\tau) = \: & \mathbb{E}_{\tau \sim \pi_{\theta}, y \sim p(y)} \left[ \frac{p(y \mid \tau)}{p(y)} - 1\right]. \\
    \end{split}
\end{equation*}
When $y$ is sampled from a uniform distribution, i.e., $p(y)$ is a constant, we have $I_f(y;\tau) = \: \alpha \; \mathbb{E}_{\tau \sim \pi_{\theta}, y \sim p(y)} \left[ p(y \mid \tau) - p(y) \right]$, where $\alpha = 1 / p(y)$. 
Following the derivation of Eq.~\eqref{eq:oracle-reward}, this optimization objective induces the linear reward function in Sec.~\ref{sec:linear-reward-function}.

\begin{table*}[!htbp]
\caption{\label{table:div}$f$-mutual information and the corresponding convex functions}
\centering
\begin{tabular}{lll}
  \toprule
  $f$-divergence    & $f(x)$                                       & $I_{f}(x;y)$                                                                                                                           \\
  \midrule
  Kullback–Leibler  & $x \log x$                                   & $\mathbb{E}_{p(x,y)} \log \frac{p(y\vert x)}{p(y)}$                                                                                    \\
  $\chi^2 $         & $(x-1)^{2}$                                  & $\mathbb{E}_{p(x,y)} \frac{p(y\vert x)}{p(y)} - 1$                                                                                     \\
  Total Variance    & $\frac{1}{2} \vert x-1\vert$                 & $\mathbb{E}_{p(x,y)} \frac{1}{2}\left\lvert 1 - \frac{p(y)}{p(y\vert x)} \right\rvert$                                                              \\
  Squared Hellinger & $(1-\sqrt{x})^{2}$                           & $\mathbb{E}_{p(x,y)} \left[2 -2 \sqrt{\frac{p(y)}{p(y\vert x)}}\right]$                                                                           \\
  Le Cam            & $\frac{1-x}{2x+2}$                           & $\mathbb{E}_{p(x,y)} \frac{[p(y\vert x) - p(y)]^{2}}{2 p(y\vert x) + 2 p(y)}$                                                          \\
  Jensen Shannon    & $x \log \frac{2x}{x+1} + \log \frac{2}{x+1}$ & $\mathbb{E}_{p(x,y)} \left[\log \frac{2 p(y\vert x)}{p(y\vert x)+p(y)} + \frac{p(y)}{p(y\vert x)} \log \frac{2 p(y)}{p(y\vert x) + p(y)}\right]$ \\
  Reverse KL        & $- \log x$                                   & $\mathbb{E}_{p(x,y)} \left[\frac{p(y)}{p(y\vert x)} \log \frac{p(y)}{p(y\vert x)}\right]$                                                         \\
  \bottomrule
\end{tabular}
\end{table*}

\section{Discriminator Noise Visualization}
\label{sec:discriminator-noise-visulization}
\begin{figure}[ht]
  \centering%
  \begin{subfigure}[ht]{0.48 \linewidth}
    \includegraphics[width=\textwidth]{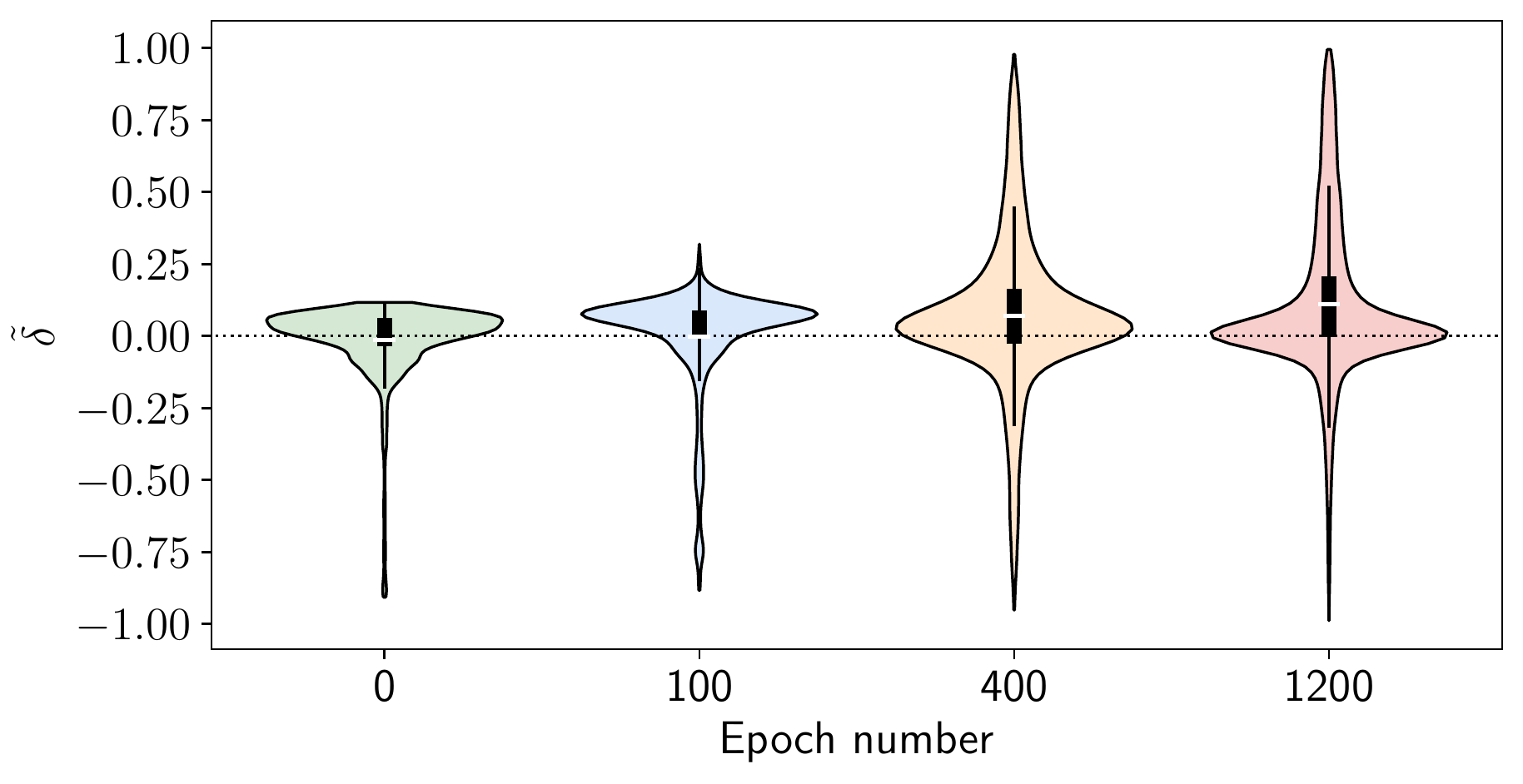}
    \caption{Accuracy-based reward}
    \label{fig:noisy-visualization-acc}
  \end{subfigure}
  \hfill
  \begin{subfigure}[ht]{0.48 \linewidth}
    \includegraphics[width=\textwidth]{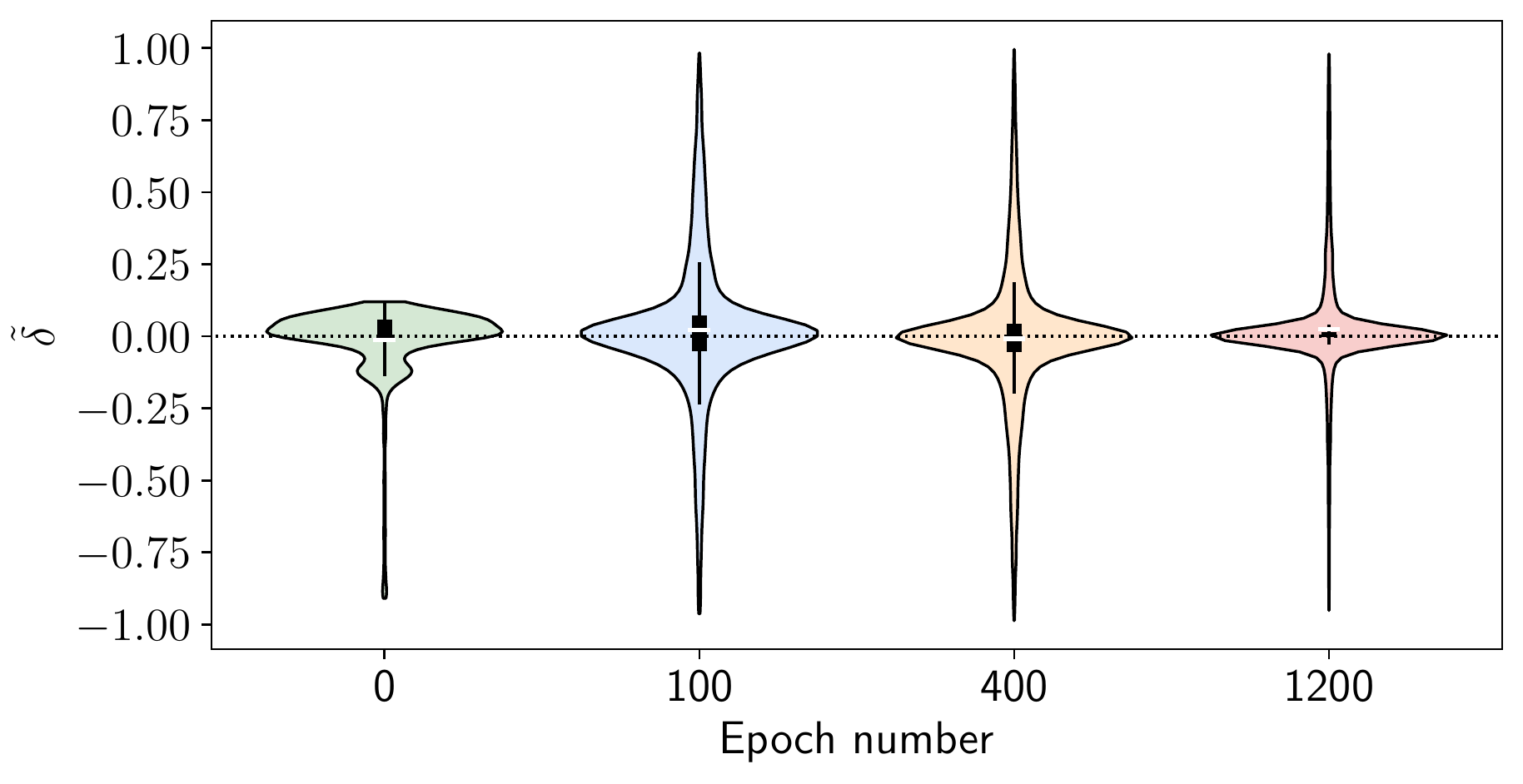}
    \caption{Clipped linear reward}
    \label{fig:noisy-visualization-linear}
  \end{subfigure}
  \caption{Visualization of the discriminator noise of RAM trained using the accuracy-based (cf. Eq.~\eqref{eq:accuracy_based_reward}) and the clipped linear reward (cf. Eq.~\eqref{eq:linear_reward_clipped}).}
  \label{fig:noisy-visualization-acc-and-linear}
\end{figure}

Besides the visualization of the discriminator noise when the policy is trained using the logarithmic reward (see Fig.~\ref{fig:noisy-visualization-log}), we also visualize the discriminator noise when using the accuracy-based and the clipped linear reward in Fig.~\ref{fig:noisy-visualization-acc-and-linear}. We can see that the discriminator noise when using the clipped linear reward has a smaller bias and variance.

\begin{figure}
  \centering%
  \includegraphics[width=0.38 \textwidth]{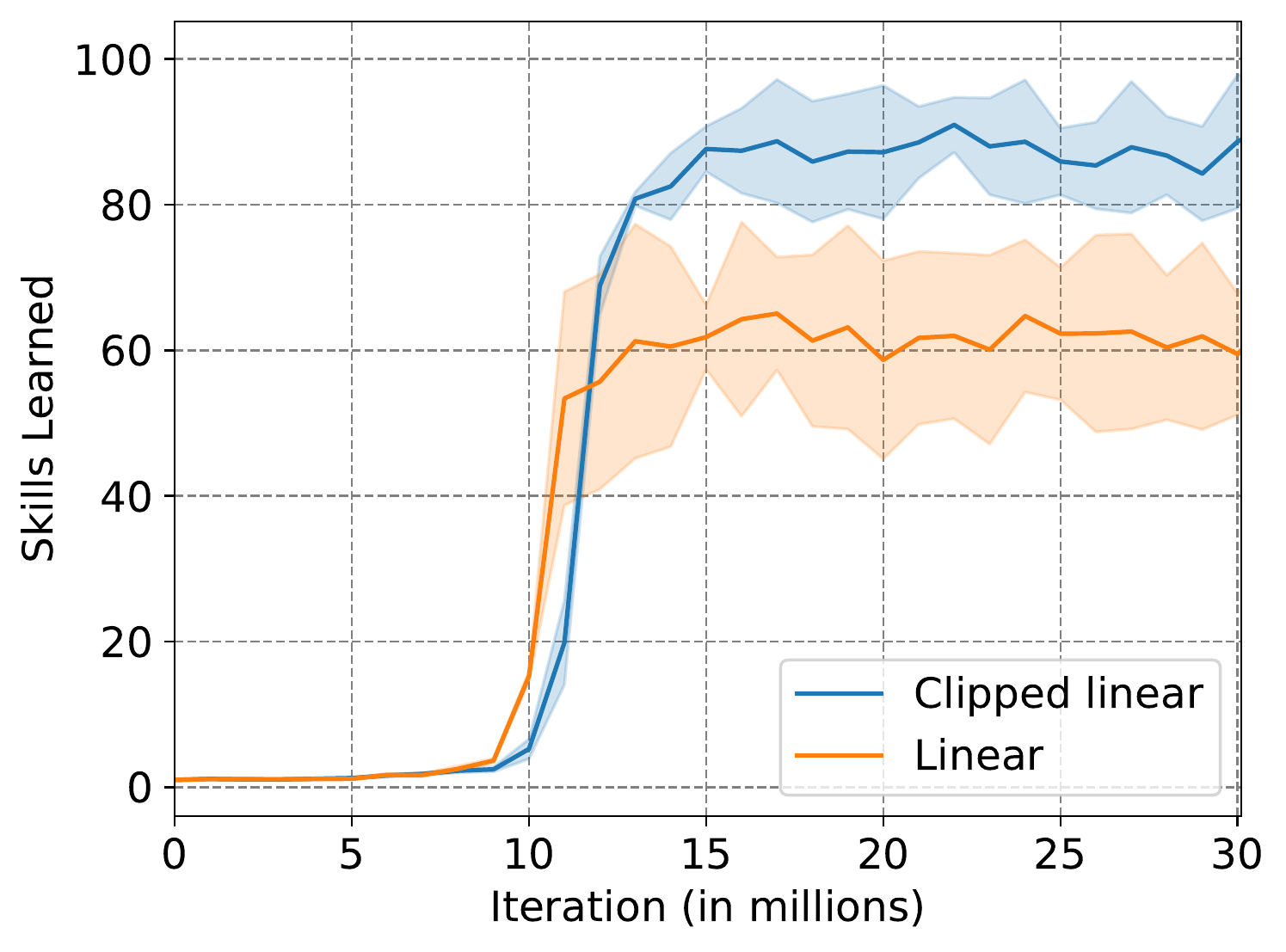}  \caption{Reward clipping on the unsupervised skill discovery task.} 
  \label{fig:skill-discovery-reward-clipping}
\end{figure}

\ml{
\section{Reward Clipping} \label{sec:reward-clipping}
We compare the performance of models trained using the logarithmic and the linear reward with and without reward clipping.
Experimental results are that the clipped logarithmic reward achieves almost the same performance on the digit recognition task on both RAM and DT-RAM, slightly better performance on the skill discovery task ($\sim$ 1.5 more learned skills), and slightly worse performance ($\sim 3.5\%$ lower accuracy) on the object counting task.
The clipped linear reward achieves almost the same performance on the object counting task, slightly better performance ($\sim 1\%$ and $\sim 1.5\%$ higher accuracy on RAM and DT-RAM respectively) on the digit recognition task, and considerable improvement ($\sim 23$ more learned skills) on the unsupervised skill discovery task (see Fig.~\ref{fig:skill-discovery-reward-clipping}).
These results suggest that reward clipping is a generally beneficial technique, which is consistent with our theoretical analysis in Sec.~\ref{sec:clipped-linear-reward-function}.
}

\section{Evaluation of Various $g$ Functions}
\label{sec:evaluation-g-functions}
\begin{figure}[ht]
    \centering
    \begin{minipage}[t]{0.45\textwidth}
        \centering
        \includegraphics[width=0.85 \textwidth]{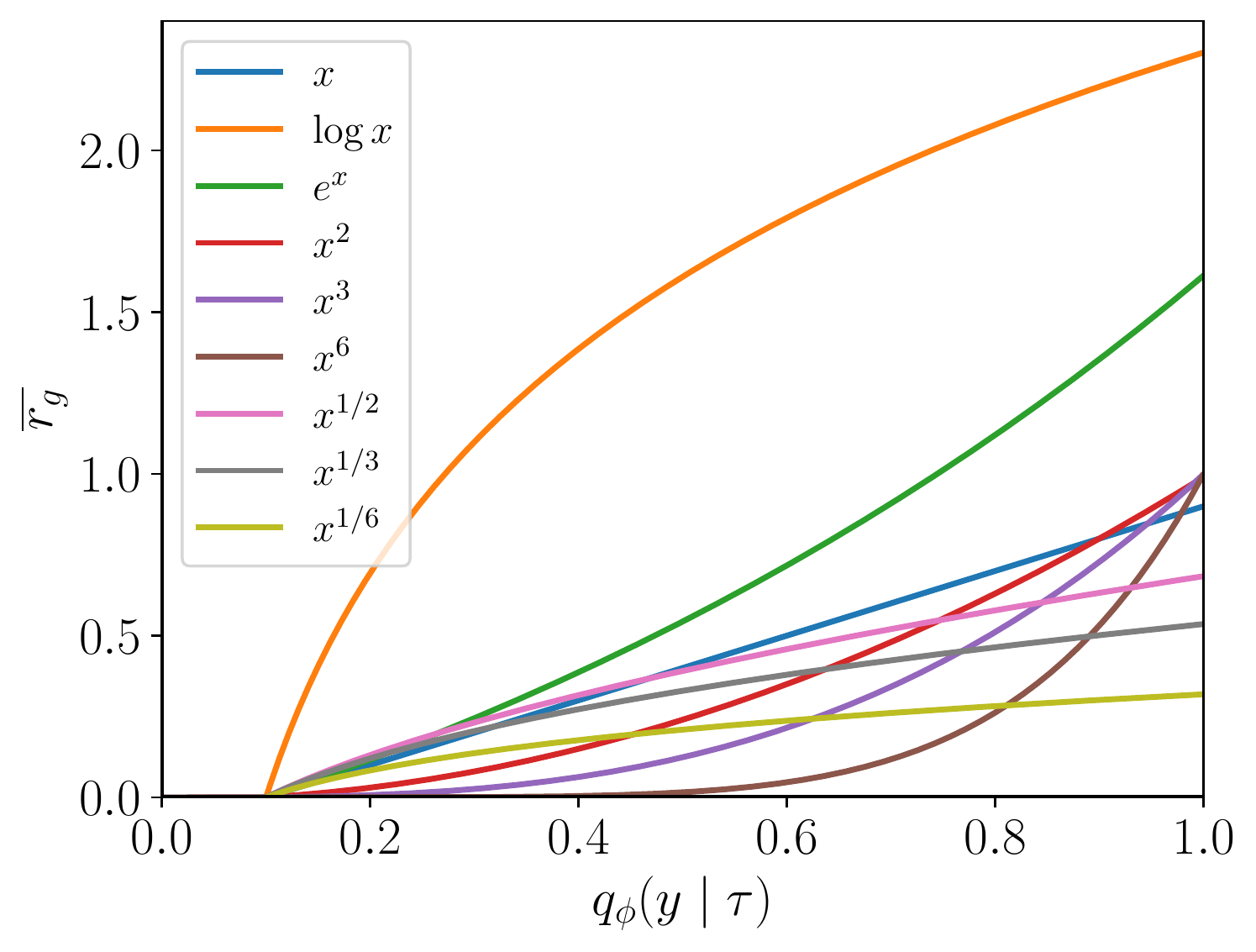}
        \caption{Clipped generalized reward with respect to the estimated posterior probability. }
        \label{fig:g-functions}
    \end{minipage} \hfill
    \begin{minipage}[t]{0.45\textwidth}
        \centering
        \includegraphics[width=0.85 \textwidth]{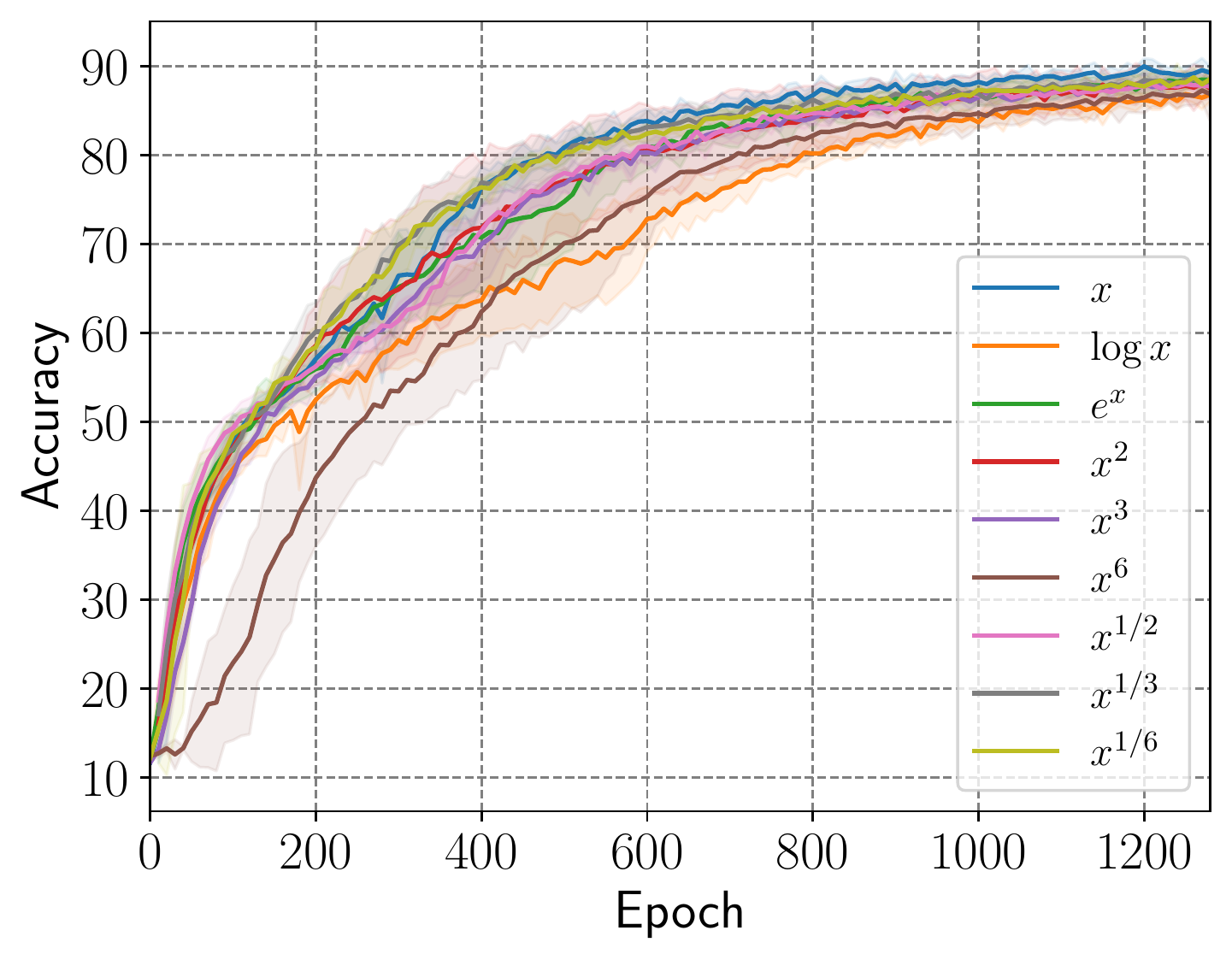}
        \caption{Evaluation of various $g$ functions using the RAM model on the digit recognition task. }
        \label{fig:digit-recognition-g-functions-no-reward-hacking}
    \end{minipage}%
\end{figure}

We evaluate several other $g$ functions in addition to the linear and logarithmic functions using the RAM model on the digit recognition task. 
Fig.~\ref{fig:g-functions} illustrates the clipped generalized reward with respect to the estimated posterior probability when using different $g$ functions (cf.~Eq.~\eqref{eq:r-g-function}). 
The reward is clipped at $q_\phi(y \mid \tau) = p(y) = 0.1$ (cf.~Eq.~\eqref{eq:linear_reward_clipped}). 
Fig.~\ref{fig:digit-recognition-g-functions-no-reward-hacking} shows training curves when using different $g$ functions. 
We can see that the linear function results in the best performance, and $g$ functions that are similar in shape to the linear function generally perform well.
The logarithmic function and function $g(x) = x^6$ perform worse than others, which can be explained from the perspective of the requirements of $g$ functions. 
Though the logarithmic function works ideally in information transmission in theory where noise is not an issue, it suffers from noisy rewards as discussed in Sec.~\ref{sec:noise-moderation}. 
Function $g(x) = x^6$, on the other hand, leads to a small bias and variance of the estimated reward, which suggests a favorable ability in noise moderation. However, it cannot transmit information with high fidelity. Its incompetence in information transmission can be observed from the shape of the corresponding plot in Fig.~\ref{fig:g-functions}, where a wide range of values, e.g., [0, 0.5], is compressed to values close to zero, leading to a substantial ignorance of information in various observations.
In contrast, the linear function achieves a trade-off between these two abilities and exhibits the best performance among all the $g$ functions considered.  

\section{Environments} \label{app:environments}
\paragraph{Cluttered-MNIST}
We generate the Cluttered MNIST dataset by a generator provided by the code repository\footnote{\url{https://github.com/skaae/recurrent-spatial-transformer-code}} of~\citeauthor{SonderbySMW15} (\citeyear{SonderbySMW15}), where we adopt the dataset configuration from~\citeauthor{mnihhgk14} (\citeyear{mnihhgk14}).
A Cluttered MNIST image is generated by randomly placing an original MNIST image (28$\times$28) and 4 randomly cropped patches (8$\times$8) from original MNIST images in an empty image (60$\times$60).
We generate 60k Cluttered MNIST images, of which $90\%$ are used for training and the rest for validation. 

\paragraph{Four-room environment}
The four-room environment is adopted from \citeauthor{Strouse2022LearningMoreSkills} (\citeyear{Strouse2022LearningMoreSkills}) and is shown in Fig.~\ref{fig:task-vis-skill-discovery}. 
There are four rooms and 104 states.
The agent is initialized at the top-left corner at each episode and can select an action from $\{ \textit{left}, \textit{right}, \textit{up}, \textit{down}, \textit{no-op} \}$ at each time step.
The length of each trajectory is 20, by which the agent is able to reach all but one state, arising the maximum number of possible learned skills 103. 
The target skill label is uniformly sampled as an integer in $\left [0,127 \right ]$ at each episode.

\paragraph{Object counting}
We create a simulation environment for the task of object counting in occlusion based on the simulation environment provided by the code repository\textsuperscript{\ref{footnote:occlusion-reasoning}} of~\citeauthor{liwklz0w21} (\citeyear{liwklz0w21}).
We use cubes of three different sizes (\emph{small}, \emph{medium}, and \emph{large}) in two different colors (\emph{red}, and \emph{blue}) as objects on the table. 
The goal object is one of the \emph{small} or \emph{medium} objects.
Each scene is initialized under the following constraints: 
1) at least one large object is on the table as an abstraction;
2) the number of other objects $N$ is sampled from a Poisson distribution ($\lambda=4$) and is clipped at a maximum number of 6;
3) one of the goal objects is occluded by an object of a larger size with a probability of $80\%$ to make occlusion happen frequently. 
The number of goal objects is uniformly sampled between 0 and $N$. 
The agent is initialized in front of the table and takes as input an egocentric RGB image with a resolution of 256$\times$256 (cf.~Fig.~\ref{fig:task-vis-object-counting}). 
The agent has three discrete actions: \textit{rotate\_right}, \textit{rotate\_left}, and \textit{stop}. 
The agent circles around the table by 30 degrees with each rotation action. The maximum number of movement steps is 6, by which the agent can move to the opposite of its initial position.
We generate offline datasets for training (100k scenes) and evaluation (1k scenes) because online occlusion checking including scene initialization in the CoppeliaSim simulator is slow.

\section{Implementation} \label{app:implementation}
\paragraph{RAM}
We use an existing implementation of the original RAM model\footnote{\url{https://github.com/kevinzakka/recurrent-visual-attention}}. 
Given an image and the coordinate of the glimpse, a glimpse network extracts visual representations of the attended patch by an MLP. 
The coordinate is mapped into representations by another MLP.
The two representation vectors have the same dimensionality of 256. They are added together to get the glimpse representations. 
A simple RNN as the core network recurrently processes glimpse representations and produces hidden representations with a dimensionality of 256 at each time step. 
A policy network takes hidden representations of the core network as input to predict the location of the next glimpse.
When the maximum number of movement steps is reached, a classification network takes hidden representations of the core network as input to produce the class prediction and finalize the task. The maximum number of movement steps is 18 in our experiments. 
The original RAM uses multi-resolution glimpses at each time step for achieving higher classification accuracy. 
The glimpse of the lowest resolution can cover almost the entire image.
This setting compromises the quality of the attention policy. 
To focus on policy learning in this work, we use a single small glimpse of size $4 \times 4$ at each time step. The idea of not using multi-resolution glimpses has been used by~\citeauthor{ElsayedKL19} (\citeyear{ElsayedKL19}) for better interpretability.
In our experiments, RAM models are trained using REINFORCE~\citep{williams92} and optimized by Adam~\citep{KingmaB14} for 1500 epochs with a batch size of 128 and a learning rate of $3\text{e-}4$.

\paragraph{DT-RAM}
The DT-RAM model used in the experiments is from our own implementation.
Instead of using two separate policy networks for location prediction and task termination respectively, which is designed for curriculum learning in the original DT-RAM, we use an integrated policy network for both location prediction and task termination. 
Same as RAM, the glimpse size is $4 \times 4$, and the maximum number of movement steps is 18 for DT-RAM. 
In our experiments, DT-RAM models are trained for 1500 epochs with the same optimization configuration as RAM models.

\paragraph{Model for unsupervised skill discovery}
The implementation of the model for unsupervised skill discovery is based on the code repository\footnote{\label{footnote:skill-discovery}\url{https://github.com/deepmind/disdain}} of~\citeauthor{Strouse2022LearningMoreSkills} (\citeyear{Strouse2022LearningMoreSkills}).
In this implementation, the model uses the last state as an abstraction of the trajectory. 
The model is trained using a distributed actor-learner setup similar to R2D2~\citep{Kapturowski2019RecurrentExperienceReplay}.
The Q-value targets are computed with Peng's $Q(\lambda)$~\citep{Peng1996IncrementalMultistepQlearning} instead of $n$-step double Q-learning.
Following~\citeauthor{Strouse2022LearningMoreSkills} (\citeyear{Strouse2022LearningMoreSkills}),
performance of the agent is evaluated using the number of learned skills
\begin{equation} \label{eq:metric-learned-skills}
n_{\text{skills}} = 2^{\mathbb{E}[\log q_\phi (y \vert \tau) - \log p(y)]}, 
\end{equation}
which can be understood as the measurement of the logarithmic reward in bits. 

\paragraph{Model for object counting}
The implementation of the model for robotic object counting is based on the code repository\footnote{\label{footnote:occlusion-reasoning}\url{https://github.com/mengdi-li/robotic-occlusion-reasoning}} of~\citeauthor{liwklz0w21} (\citeyear{liwklz0w21}).
We replace the REINFORCE algorithm with PPO for more efficient training.
The implementation of the PPO algorithm is based on the code repository\footnote{\label{footnote:babyai}\url{https://github.com/mila-iqia/babyai}} of~\citeauthor{babyai_iclr19} (\citeyear{babyai_iclr19}).
The model consists of a pretrained and fixed ResNet18~\citep{HeZRS16} to extract feature maps from its \textit{conv3} layer.
The feature maps are then passed through two CNN layers and an average pooling layer to get visual representations of dimension 256.
The index of the target object is mapped into a 10-dimensional embedding, which is called the goal representation.
The visual and goal representations are concatenated together as the input of an RNN network, which recurrently produces hidden representations at each time step for the policy network and classification network.
When the policy network selects the \textit{stop} action, the classification network is triggered to produce the prediction of the number of the target object.
We train the model for 2M episodes. Five processes are used to collect experience with a horizon of 40 steps. We train the model using Adam~\cite{KingmaB14} with a learning rate of $1\text{e-}4$.
Other hyperparameters of PPO are the same as the original implementation\textsuperscript{\ref{footnote:babyai}} except that we use 10 epochs of minibatch optimization and 5 parallelization processes.

\section{DISDAIN} \label{app:disdain}
The reward of the DISDAIN method is $r = r_{\text{log}} + \lambda r_{\text{DISDAIN}}$, where $r_{\text{log}}$ is the logarithmic reward function (cf. Eq.~\eqref{eq:log_reward}), $\lambda$ is a weighting coefficient, and $r_{\text{DISDAIN}}$ is an auxiliary ensemble-based reward calculated as 
\begin{equation} \label{eq:disdain_reward}
r_{\text{DISDAIN}} = \mathbb{H} \left[ \frac{1}{N} \sum_{i=1}^N q_{\phi_i} (y \mid \tau) \right] - \frac{1}{N} \sum_{i=1}^N \mathbb{H} \left[ q_{\phi_i} (y \mid \tau) \right],
\end{equation}
where $N$ is the number of discriminators of the ensemble, and $\mathbb{H}[X]$ is the entropy of random variable $X$. The DISDAIN reward is essentially the estimation of the epistemic uncertainty of the discriminator. 


\section{Additional Results}

\subsection{Case Study} \label{app:case-study}
\subsubsection{Hard attention for digit recognition}

\begin{figure*}[ht]
  \begin{tabular}{ccc}
  \centering%
  \begin{subfigure}[t]{0.46\linewidth}
  \includegraphics[width=\textwidth]{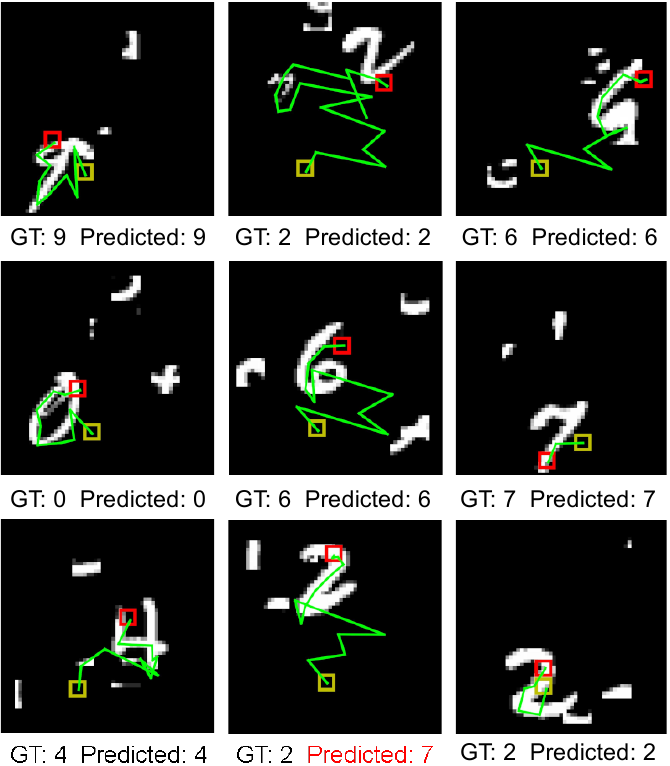}
  \caption{Clipped linear}   
  \end{subfigure} 
  & \tikz{\draw[-,black, densely dashed, thick](0,0.50) -- (0,7.95);} &
  \begin{subfigure}[t]{0.46\linewidth}
  \includegraphics[width=\textwidth]{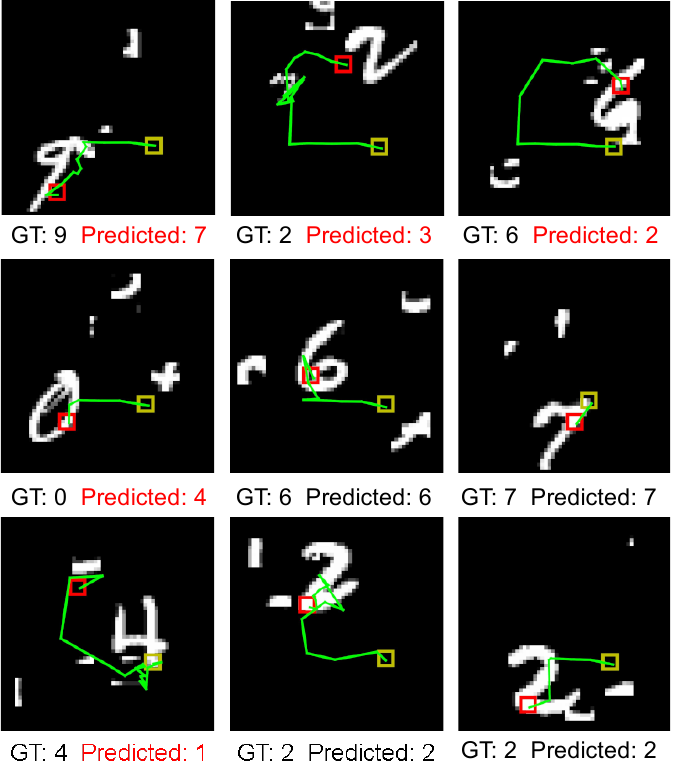}
  \caption{Clipped logarithmic}   
  \end{subfigure}
  \end{tabular}   
  
  \centering
   \begin{subfigure}[t]{0.46\linewidth}
  \includegraphics[width=\textwidth]{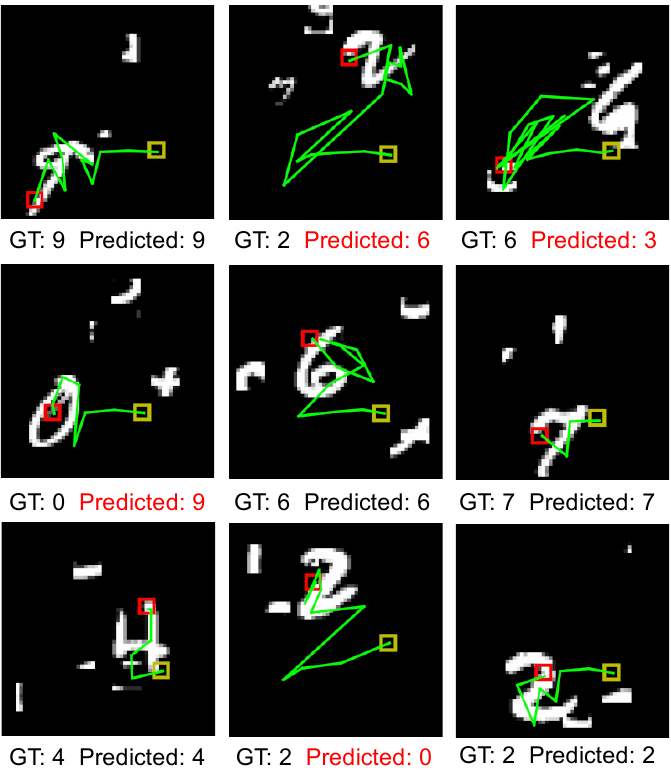}
  \caption{Accuracy-based}   
  \end{subfigure}
  \caption{Comparison of DT-RAM models trained by different reward functions. GT: the ground-truth class; Predicted: the predicted class. Red indicates incorrect predictions.}
  \label{fig:case-study-digit-recognition}
\end{figure*}

In Fig.~\ref{fig:case-study-digit-recognition}, we provide cases of the DT-RAM model on the digit recognition task for intuitive comparison between the model trained using different reward functions. 
All the cases are randomly sampled without any cherry-picking.
We can see that trajectories generated by the model trained using the clipped linear reward can cover sufficient information for recognizing the digit, while trajectories generated by the model trained using the logarithmic reward function tend to be pessimistic, e.g., trajectories in cases of digit 9 and digit 6 in the first row, digit 0 in the second row, and digit 4 in the third row.
The exploration trajectories generated by the model trained using the accuracy-based reward tend to sample less informative areas, e.g., trajectories in cases of digit 6 in the first row, and digit 2 in the third row and second column, which may account for its low accuracy. 

\subsubsection{Robotic object counting}
\begin{figure*}
    \centering
    \includegraphics[width=\linewidth]{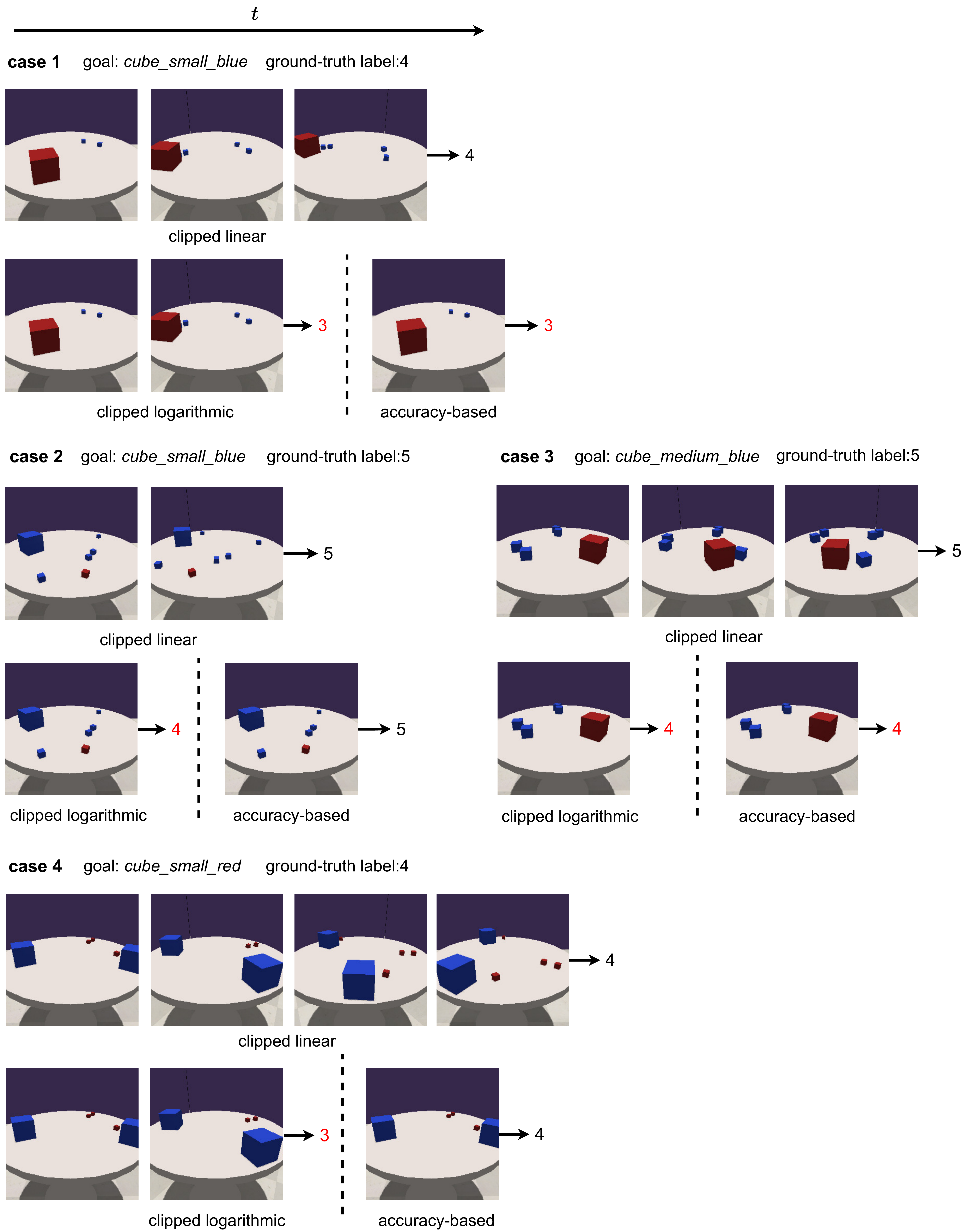}
    \caption{Comparison of models trained by different reward functions on the robotic object counting task. The number next to the arrow after a sequence of egocentric views is the number of goal objects predicted by the agent. Red numbers indicate wrong predictions. }
    \label{fig:object-counting-cases}
\end{figure*}

Fig.~\ref{fig:object-counting-cases} shows examples of the pessimistic exploration issue when using the logarithmic and the accuracy-based reward function. The agent trained using the accuracy-based reward function chooses not to move, and the agent trained using the logarithmic reward function terminates exploration too early to acquire sufficient information for predicting the number of the target object. 
They guess the number of target objects based on insufficient observations, while the agent trained using the clipped linear reward function learns to choose a reasonable number of movement steps to explore the environment. 

\subsection{State Occupancy in Unsupervised Skill Discovery} \label{app:state-occupancy}
\begin{figure*}[ht]
  \centering%
  \begin{subfigure}[t]{0.7\linewidth}
  \includegraphics[width=\textwidth]{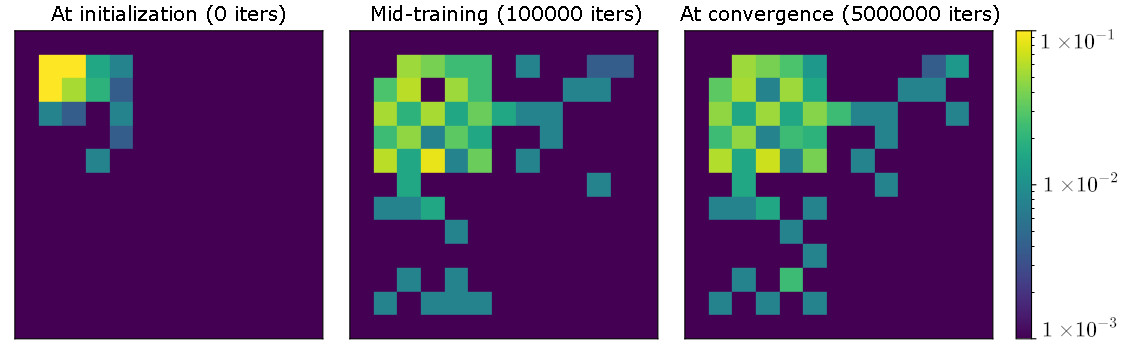}
  \caption{Clipped logarithmic}   
  \end{subfigure}

  \begin{subfigure}[t]{0.7\linewidth}
  \includegraphics[width=\textwidth]{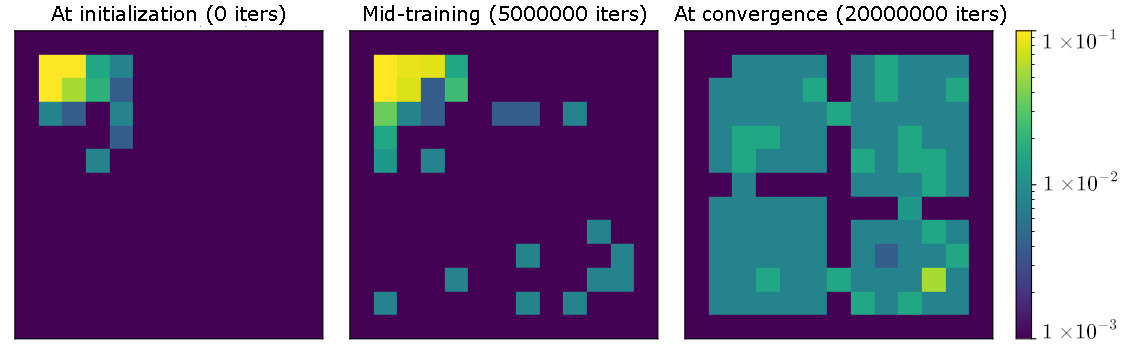}
  \caption{DISDAIN}   
  \end{subfigure} 
  
  \begin{subfigure}[t]{0.7\linewidth}
  \includegraphics[width=\textwidth]{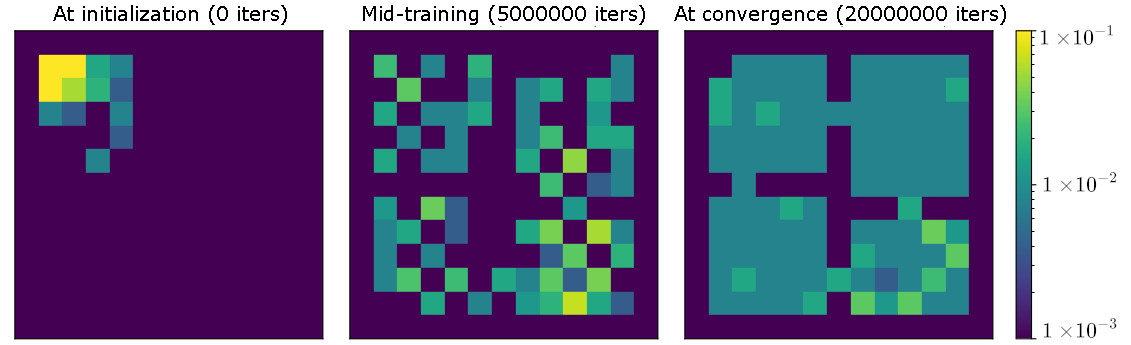}
  \caption{Clipped linear}   
  \end{subfigure} 

  \caption{States reached using different reward functions. Plots depict ratios of final states reached after performing 10 trajectories per skill. 
  The ratio is clipped between 0.001 and 0.1 for the sake of visualization. Note that the number of iterations at convergence is different (see Fig.~\ref{fig:skill-discovery-baseline-comparison}).
  }
  \label{fig:skill-discovery-state-occupancy}
\end{figure*}

Fig.~\ref{fig:skill-discovery-state-occupancy} demonstrates state occupancy reached using different reward functions at initialization, at the intermediate stage, and at convergence during training. 
We can see that using the clipped linear reward function, the agent learns to reach all states as using the DISDAIN reward, while the agent mainly explores the first room when using the clipped logarithmic reward function. 

\end{document}